\pdfoutput=1
\documentclass[11pt]{article}

\usepackage[final]{acl}

\usepackage{times}
\usepackage{latexsym}

\usepackage[T1]{fontenc}
\usepackage[utf8]{inputenc}

\usepackage{microtype}

\usepackage{inconsolata}

\usepackage{color}
\usepackage{xcolor,colortbl}

\definecolor{Gray}{gray}{0.15}
\definecolor{LightCyan}{rgb}{0.9,1,1}
\usepackage{booktabs}
\usepackage{graphicx}
\usepackage{multirow}
\usepackage{amsmath}
\usepackage[normalem]{ulem}
\usepackage{bbm}
\usepackage{enumitem}

\newcommand{\todo}{\textcolor{black}}

\title{Selective Annotation via Data Allocation: These Data Should Be Triaged to Experts for Annotation Rather Than the Model}

\author{
Chen Huang$^{\spadesuit\heartsuit}$, \quad
Yang Deng$^{\clubsuit}$, 
\quad 
\textbf{Wenqiang Lei}$^{\spadesuit\heartsuit}$\thanks{Corresponding author.}, \quad
\textbf{Jiancheng Lv}$^{\spadesuit\heartsuit}$, \quad
\textbf{Ido Dagan}$^{\diamondsuit}$
\\
${\spadesuit}$ Sichuan University \quad ${\clubsuit}$ Singapore Management University \quad ${\diamondsuit}$ Bar-Ilan University\\
${\heartsuit}$ Engineering Research Center of Machine Learning and Industry Intelligence, Ministry of Education, China \\
huangc.scu@gmail.com \quad ydeng@smu.edu.sg \\ \{wenqianglei, lvjiancheng\}@scu.edu.cn \quad dagan@cs.biu.ac.il
}

\begin{document}
\maketitle
\begin{abstract}
To obtain high-quality annotations under limited budget, semi-automatic annotation methods are commonly used, where a portion of the data is annotated by experts and a model is then trained to complete the annotations for the remaining data. However, these methods mainly focus on selecting informative data for expert annotations to improve the model predictive ability (i.e., triage-to-human data), while the rest of the data is indiscriminately assigned to model annotation (i.e., triage-to-model data). This may lead to inefficiencies in budget allocation for annotations, as easy data that the model could accurately annotate may be unnecessarily assigned to the expert, and hard data may be misclassified by the model. As a result, the overall annotation quality may be compromised. To address this issue, we propose a selective annotation framework called SANT. It effectively takes advantage of both the triage-to-human and triage-to-model data through the proposed error-aware triage and bi-weighting mechanisms. As such, informative or hard data is assigned to the expert for annotation, while easy data is handled by the model. Experimental results show that SANT consistently outperforms other baselines, leading to higher-quality annotation through its proper allocation of data to both expert and model workers. We provide pioneering work on data annotation within budget constraints, establishing a landmark for future triage-based annotation studies.
\end{abstract}

\section{Introduction}
\begin{figure}[tb]
   \centering  
    \setlength{\abovecaptionskip}{2pt}   
    \setlength{\belowcaptionskip}{2pt}
   \includegraphics[width=0.47\textwidth]{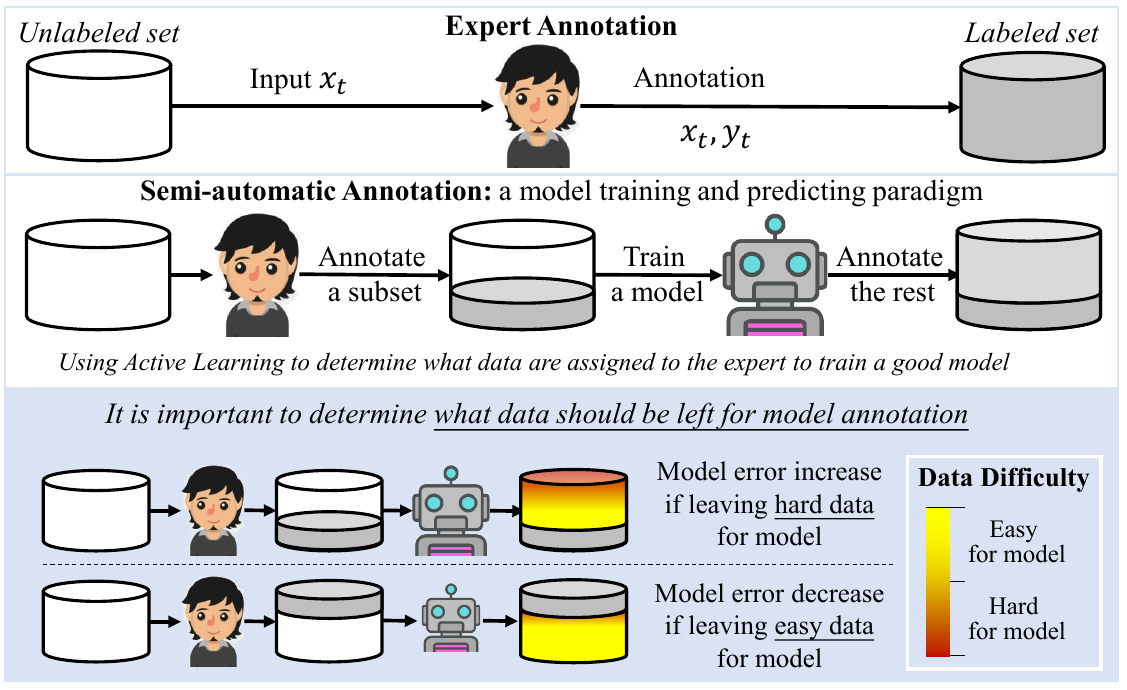} 
   \caption{Semi-automatic annotation focuses on triage-to-human data, overlooking triage-to-model data.} 
   \vspace{-7mm}
   \label{demograph1}
\end{figure}

Creating a high-quality and fully annotated dataset by human experts is criticized as being expensive. 
%Obtaining such a dataset
It is impossible to complete the annotation if we are operating on a limited budget \cite{chen-etal-2021-language-resource}, especially when dealing with large-scale data corpora on the web~\cite{hedderich2021anea, feng-lapata-2008-automatic}. %datasets
% \dy{data corpora on the web} 
In this paper, \textbf{we study the data annotation problem within the constraints of limited budgets}, where the goal is to achieve high-quality annotations despite having insufficient budgets to hire experts for annotating the entire dataset. 
To tackle this challenge, a typical solution is to conduct semi-automatic annotation \cite{chen2020survey, yao2023labels} (cf. Fig. \ref{demograph1} (\textit{middle})). 
It typically allocates the budget to hire experts for annotating a portion of the data, and then utilizes the annotated data to train a prediction model for completing the annotations for the remaining data. 
The choice of the model can vary, ranging from sophisticated models with large parameters \cite{bryant-etal-2017-automatic, schulz2019analysis, hedderich2021anea} to lightweight models \cite{chen2020jit2r, pmlr-v133-desmond21a} depending on the available hardware budgets. 
However, such methods often overlook the data allocation problem, \textit{i.e.}, the process of deciding which data should be annotated by human experts and which should be left for the prediction models to annotate. 
%This can result in a waste of annotation budgets because some data that the model could have annotated correctly may be given to the expert for annotation. Consequently, the utilization of annotation budgets is not optimized. 
This may lead to inefficiencies in budget allocation for annotations, as certain data that the model could accurately annotate is unnecessarily assigned to experts. Consequently, the overall annotation quality within the same budget may be compromised.
% Consequently, this hinders the optimization of annotation quality under limited budgets. 
% CHEN: I removed the term 'long-term'. Seems difficult to explain?

% Given a fixed amount of unlabeled data, it’s essential to optimally allocate annotations among the expert and the model to achieve the better annotation quality

In the light of this challenge, \textbf{we take the first step to re-formulate the semi-automatic annotation under limited budgets as a data triage problem}, which aims to optimize the annotation quality under limited budgets by determining the allocation of data to human annotator (namely \textbf{triage-to-human data}) and data to be assigned for model prediction (namely \textbf{triage-to-model data}). 
Existing semi-automatic annotation methods \cite{pmlr-v133-desmond21a, su2022selective, li2023coannotating} mainly focus on selecting unlabeled data for expert annotations to improve the model predictive ability (\textit{i.e.}, triage-to-human data) through random strategies or active learning (AL), while the rest of data is indiscriminately assigned to model annotation (\textit{i.e.}, triage-to-model data). 
However, as shown in Fig. \ref{demograph1} (\textit{bottom}), even if we constantly improve the predictive ability via triage-to-human data, the model still inevitably makes errors in its annotations.
Not to mention that the data allocation strategies of existing methods may not effectively improve the model predictive ability \cite{tang2019self, mindermann2022prioritized}. 
To address this issue, determining what data should be assigned for model annotation (triage-to-model data) is crucial. 
A model could achieve high-quality annotations if assigned with the data that is easy for model to predict, while experts annotate the hard ones. 
Therefore, when solving the data triage challenge, it's important to take both triage-to-human and triage-to-model data into account. 
In this way, the budget utilization could be improved through efficiently allocating both human and model annotations.

To this end, we propose a novel \uline{S}elective \uline{AN}no\uline{T}ation framework (SANT), which optimizes the utilization of limited annotation budgets by balancing the importance of triage-to-human and triage-to-model data.  
Firstly, an AL-based mechanism is deployed to assign higher weights to data that help improve the model predictive ability. 
Secondly, an \uline{E}rror-\uline{A}ware \uline{T}riage (EAT) mechanism is proposed to assess the "hardness" weight of data, with the aim of redirecting the focus toward triage-to-model data. In specific, 
EAT dynamically estimates the probability of model error using triage-to-human data from previous rounds. 
Finally, a bi-weighting mechanism is adopted to aggregate different advantages of prioritizing triage-to-human and triage-to-model data. 
Such mechanism adaptively adjusts the importance of data so that informative or hard-to-predict data is assigned to humans for annotation. 
As such, the annotation budget are efficiently utilized to secure the model predictive ability and to leverage the strengths of the expert on hard data annotation simultaneously.
% In this manner, the data annotation is successfully formulated as a data triage problem by revisiting the roles of humans and models in data annotation. % clarify the work duty of the expert and the model.
% and that the model is trained properly

Our experiment results show that SANT consistently improves annotation quality across various annotation budgets, surpassing the best semi-automatic annotation method by an average of +0.50\%, +4.86\%, and +4.54\% on the three annotation tasks, respectively. 
Recently, as large language models (LLMs) possess strong zero-shot and few-shot capabilities for data annotation \cite{he2023annollm, gilardi2023chatgpt, ding-etal-2023-gpt}, we further adopt ChatGPT as a strong automatic annotation baseline. Notably, SANT exhibits a better annotation quality over ChatGPT, even with the aid of few-shot in-context learning and Chain-of-Thought (CoT) prompting \cite{brown2020language, wei2022chain}. 
These results highlight that proper allocation of data to both human and model workers results in higher-quality annotation. 
% Moreover, our in-depth analysis shows that SANT properly allocates triage-to-human data to train the model and enhance its predictive capability. Additionally, it allocates triage-to-model data to improve the model annotation quality by providing easy-to-predict for the model. 
Further analysis highlights the importance of the triage-to-model data, which are more beneficial to obtain higher-quality annotations compared to the triage-to-human ones. 
In conclusion, we pioneer work on data annotation with limited budgets by formulating a data triage problem. We set a landmark for future triage-based annotation methods. 
Our contributions are as follows:
\begin{itemize}[leftmargin=*, itemsep=-4pt]
%    \item Calling attention to the annotation quality under the limited budget, we study the roles of both triage-to-human and triage-to-model data. We verify the importance of determining what data should be left for model annotation, which is largely overlooked by existing studies.
    \item We present the first work to study the roles of both triage-to-human and triage-to-model data. We verify the importance of triage-to-model data, which is largely overlooked by existing studies.
%     \item We take the first step in re-formulating the semi-automatic annotation as a data triage problem, and propose a selective annotation framework, called SANT. It considers triage-to-human and triage-to-model data simultaneously by introducing two novel mechanisms to optimize the utilization of limited annotation budgets directly.
    \item We re-formulate semi-automatic annotation as data triage and propose a selective annotation framework, achieved by optimizing two novel mechanisms (i.e., EAT and bi-weighting mechanisms) for utilizing limited budgets.
    \item We show that proper allocation of data to both the expert and model workers results in higher-quality annotation. SANT effectively balances the advantages of two factors: improving model predictive ability (i.e., \textit{triage-to-human data}) and triaging data based on the model's predictive ability (\textit{triage-to-model data}).
    % , varies at different stages of data annotation
    % The analysis on the superiority of SANT over other triage strategies also provide insights for future studies on quality-resource trade-off.
\end{itemize}

% uncertainty-based triage need to set a threshold for rejecting action. \cite{cortes2016learning, geifman2019selectivenet} tunes the threshold over a set of values.

\section{Related Work}
\label{dddd8374}
% We outline the differences between our work and existing data annotation studies below. 
% Further literature reviews, including those on active learning techniques, can be found in the Appendix \ref{morerelate}.

\textbf{Data Annotation}. Human annotations are known to be expensive \cite{chen-etal-2021-language-resource, hedderich2021anea, feng-lapata-2008-automatic}. 
To this end, many studies have explored human-machine cooperative data annotation \cite{lu2024does, li2023coannotating, ding2024data, huang-etal-2024-araida, huang2024comatchinghumanmachinecollaborativelegal}. 
In this case, a portion of the data is designated for human annotation (triage-to-human data), while the remaining data is allocated for model annotation (triage-to-model data). To further enhance the model’s predictive capabilities, existing methods are often formulated as a "model training and prediction" paradigm, where the model iteratively learns from the annotated data and then makes predictions. \cite{dalvi2016ike,li2015vinery, bryant-etal-2017-automatic, chen2020survey, wang2023chatgpt, tan2023evaluation}. Considering the hardware budgets and time efficiency requirements, light-weighted models \cite{pmlr-v133-desmond21a, chen2020jit2r}, such as label propagation \cite{NIPS2003_87682805} and MLP are preferred. Recently, uncertainty-based methods have been used to select a data subset for human annotations \cite{zhang2021human, pmlr-v133-desmond21a, su2022selective, ein-dor-etal-2020-active}, improving the overall annotation quality by refining the model using more informative data.
However, current methods concentrate solely on selecting triage-to-human data, while neglecting the importance of triage-to-model data. This fails to optimize data allocation effectively. In contrast, SANT considers both types of data. 
%, thus it achieves high-quality data annotations based on limited annotation budgets.

\textbf{Active learning}. Current data allocation practices in the field of data annotation draw inspiration from the active learning \cite{Ren2020ASO}, where data with high uncertainty are prioritized for human annotation. This aligns with active learning's preference for informative or representative data \cite{desmond2021semi, loquercio2020general}. To our knowledge, no existing research has explored the application of other active learning techniques for data allocation in annotation. Our work, SANT, extends this approach by introducing a novel perspective. It prioritizes triage-to-model data, essentially introducing the anti-current learning \cite{braun2017curriculum} into the task of data allocation (i.e., we assign hard-to-predict data for human annotation). SANT seamlessly integrates active learning algorithms with the anti-current learning approach through our proposed bi-weighting mechanism. Our experiments demonstrate the effectiveness of this novel framework in enhancing the quality of data annotation.
%\begin{figure}[t]
%   \centering  
%   \includegraphics[width=0.47\textwidth]{fig/f2345.pdf} 
%   \caption{Example of SANT's working pipeline. The expert and the model complete the data annotation cooperatively and iteratively.}  
   %\vspace{-3mm}
   %\label{demograph}
%\end{figure}  

\section{SANT: Selective Annotation}
We introduce the proposed Selective Annotation (SANT) framework.
In Section \ref{overd}, we provide the problem definition of the data annotation process, while Section \ref{mechanism} introduces the proposed framework SANT and two novel mechanisms.
In Appendix \ref{syenx}, we introduce the learning details of our framework. 
Following previous works on semi-automatic annotation \cite{chen2020jit2r, pmlr-v133-desmond21a, hedderich2021anea} and human-model interaction \cite{hwa2000sample, kristjansson2004interactive, huang-etal-2023-reduce}, 
we assume the cost required for human annotation is constant for each data point, and estimate the annotation budgets using the number of data points.

\subsection{Problem Definition}
\label{overd}
% \dy{We reformulate the data annotation under limited budgets as a data triage problem, where a data triage module $d$ assigns data to either the expert or the model $f$ on-the-fly. }
We reformulate the data annotation under limited budgets as a data triage problem, where a data triage module $d$ assigns data to either the expert or the model $f$ on-the-fly.
%SANT follows a similar sequential annotation process as the semi-automatic annotation paradigm, where a data triage module $d$ assigns data to either the expert or the model $f$ on-the-fly. 
The triage module outputs a triage signal of either $0$ or $1$ to delegate the annotation duty. If the signal is $0$, the model is responsible for data annotation. Conversely, if the signal is $1$, the human is assigned, and the model is updated using previousl human-annotated data. 
% In semi-automatic annotation, the triage module utilizes an AL method or random strategy. However, SANT enhances the triage module by incorporating the proposed two mechanisms (i.e., EAT and bi-weighting mechanisms). 
%\dy{merge these two sentences to the next subsection 3.2}

% In line with the \textit{learning to defer} research \cite{raghu2019algorithmic, okati2021differentiable}, SANT is composed of two parts, i.e. an off-the-shelf model annotator $f$ and a data triage module $d$.  

The annotation process terminates when one of two conditions is met: either the limited budgets are exhausted or all the data is annotated. In the first case, if the budgets are depleted before completing the entire dataset, the model annotator must continue with the remaining annotations. In the second case, it is possible for all data to be annotated before the budgets are exhausted if the amount of data assigned to the human during process is less than the budgets. To maximize the utilization of the human budget, a portion of the model-annotated data, selected by the customized data triage strategy, is re-allocated to the human annotator until their budget is depleted.

\subsection{Proposed Framework}
\label{mechanism}
Semi-automatic annotation methods use active learning (AL) in the data triage module, which primarily prioritize triage-to-human data but overlook the triage-to-model data. As a result, they fail to efficiently and directly optimize budget utilization. 
%To address this limitation, SANT goes a step further than AL by introducing two innovative mechanisms, namely EAT and bi-weighting mechanisms. These mechanisms consider both human and model-based triage data simultaneously, leading to more effective budget optimization. 
To this end, SANT introduces two innovative mechanisms, namely EAT and bi-weighting mechanisms, which consider both human and model-based triage data for budget optimization. 

\subsubsection{AL-based mechanism} 
Semi-automatic annotation and SANT are flexible to any off-the-shelf AL methods\footnote{The general preference is for stream-based AL methods over pool-based ones, as they tend to be more time-efficient.}, such as, leveraging the uncertainty of the model prediction \cite{desmond2021semi} or using a differentiable neural network \cite{wang2022boosting}. Formally speaking, given data $x_t$ and the model $f_t$ at round $t$, $x_t$ is assigned to the expert if the AL-based mechanism $d_t^{AL}$ gives $x_t$ a large weight. If the AL-based mechanism is implemented using a learnable neural network, both AL and $f_t$ are jointly optimized/updated, otherwise only $f_t$ is updated. Note that the AL-based mechanism solely focuses on triage-to-human data, which limits its ability to optimize budget utilization directly and efficiently.

\subsubsection{Error-aware triage mechanism (EAT)}
\label{asduionweasdo}
Unlike existing AL-based semi-automatic annotation, we propose EAT to prioritize triage-to-model data by assigning hard-to-predict data to the expert and reserving easy-to-predict data to the model. Specifically, EAT assigns data $x_t$ to the expert if the EAT-based mechanism $d_t^{EAT}$ gives $x_t$ a high weight, which is determined by the probability of $f_t$ misclassifying $x_t$. The model $f_t$ and EAT are then jointly optimized/updated. If $d_t^{EAT}$ does not assign a high weight to $x_t$, then the model $f_t$ acts as the annotator without requiring human involvement. Notably, EAT can be parameterized using any neural network, depending on the available hardware budgets. Here, EAT's input and learning objectives are presented below.

\textit{Input of EAT}. Instead of learning from training data independently \cite{mozannar2020consistent, wilder2021learning, raghu2019algorithmic, okati2021differentiable}, to enhance EAT's input features, we propose capturing the relationship between input data and its neighboring data from the same batch. This includes the use of weighted neighborhood entropy, which measures the model's predictive uncertainty on the local area of input data $x_t$. To calculate this entropy, we retrieve the top-$k$ neighbors $N_t(x_t)$ and calculate the cosine similarity between $x_t$ and each neighbor $x_i \in N_t$. We store these similarity values in vector $C_t(x_t) \in R^{k}$. Then, we calculate the entropy of the model prediction $f_t(x_i)$ for each $x_i \in N_t(x_t)$ and store them in vector $E_t(x_t) \in R^{k}$. Finally, we obtain the weighted neighborhood entropy of $x_t$ by using the element-wise multiplication operator $\odot$ on vectors $E_t(x_t)$ and $C_t(x_t)$. The input of EAT is then the concatenation of the embedding of unlabeled data $x_t$, the model output $f_t(x_t)$, and the weighted neighborhood entropy. 

\textit{Learning objective of EAT}. 
EAT is designed to allocate hard-to-predict data, which are beyond the model's predictive ability, to the expert. To achieve this, EAT estimates the likelihood of the model making errors on unlabeled data $x_t$.
% To ensure that hard-to-predict data that are beyond the model's predictive ability are allocated to the expert, EAT aims to estimate the probability of the model making mistakes on unlabeled data $x_t$. 
Specifically, EAT is required to minimize the following loss.
% EAT aims to allocate hard data that are beyond the model predictive ability to the expert. In this way, model annotation quality on the triage-to-model data should be higher than the quality on the triage-to-human data. To achieve this, EAT is required to estimate the probability of the model making mistakes on unlabeled data $x_t$. To this end, we minimize the following loss.
\begin{equation}
\label{dyue84nu}
    \small
    L_d = \sum_{\{x_t,y_t\}} \ell_d(\mathbbm{1}[y_t \neq \max(f_t(x_t))], d_t^{EAT}(x_t)).
\end{equation}
Here, $\mathbbm{1}[\cdot]$ is an indicator function. $y_t \in R$ and $f_t(x_t) \in R^{\mathbf{C}}$ are the ground truth from the human and predicted outputs of the model, where $\mathbf{C}$ is the number of classes. Also, the operator $\max(\cdot)$ outputs the index of the maximal value of the input vector. In our experiments, $\ell_d$ comes in the form of the negative log-likelihood or binary cross-entropy loss, depending on the annotation tasks. Moreover, We further follow recent works on few-shot learning \cite{wang2018large, li2020boosting} and utilize the max-margin loss to improve SANT's ability to learn and generalize from limited labeled data on the fly. Apart from the aforementioned loss $L_d$, we add the following max-margin loss $L_m$ (See Appendix \ref{sensi} for ablation experiments on $L_m$).
\begin{equation}
\small
\begin{split}
    L_m &= \max(0, \alpha + \overline{L_f^{m}} - \overline{L_f^{h}}),\\
    \overline{L_f^{m}} &= \frac{\sum_{\{x_t,y_t\}}\mathbbm{1}[d_t^{EAT}(x_t)< 0.5]\ell_f(x_t, y_t)}{\sum_{\{x_t,y_t\}}\mathbbm{1}[d_t^{EAT}(x_t)< 0.5]},\\
    \overline{L_f^{h}} &= \frac{\sum_{\{x_t,y_t\}}\mathbbm{1}[d_t^{EAT}(x_t) \geq 0.5]\ell_f(x_t, y_t)}{\sum_{\{x_t,y_t\}}\mathbbm{1}[d_t^{EAT}(x_t) \geq 0.5]}.
\end{split}
\end{equation}
Here, $\alpha>0$ is a margin hyper-parameter.  Given the fixed model, EAT carefully allocate the expert and the model annotations so that it ensures that the average loss of the model on the triage-to-human data $\overline{L_f^{h}}$ is much higher than the triage-to-model data $\overline{L_f^{m}}$. Finally, the learning objective of EAT is given by $L_{EAT}=L_d + L_m$.

\subsubsection{Bi-weighting mechanism}
\label{7ujm0ok}
The AL-based mechanism focuses on triage-to-human data to improve the model's predictive ability. However, it may still struggle with hard data beyond the model capacity. In contrast, our proposed EAT-based mechanism prioritizes triage-to-model data and reserves hard data for human annotation, which allows the model to handle relatively easy data. However, models trained on hard data may have weaker predictive ability. To find the optimal trade-off, the bi-weighting mechanism aims to ensure that the model is trained properly and can annotate relatively easy data effectively.

% AL-based mechanism focuses solely on triage-to-human data. It triages informative data for human annotation so that the model's predictive ability may improve. However, the model may still struggle with hard data that are beyond its capacity. To this end, our proposed EAT-based mechanism prioritizes triage-to-model data and leave hard data for human annotation, the remaining data for the model may be relatively easy. However, the predictive ability of models trained on hard data may be weaker.  Therefore, the bi-weighting mechanism is to find the optimal the trade-off, ensuring the model is trained properly and able to annotate relatively easy data.

The bi-weighting mechanism aggregates the advantages of prioritizing both triage-to-human and triage-to-model data by introducing a simple yet efficient method. It scores $x_t$ using $d_t^{bi}(x_t)$, which is the product of $d_t^{AL}(x_t) \in [0,1]$ and $d_t^{EAT}(x_t) \in [0,1]$, to adjust the importance of data. This ensures that informative or hard data are assigned to humans for annotation, and that the model is trained properly. 
Considering the sample efficiency of AL is more advantageous when the model is trained based on a few labeled data, we further control the importance of the AL score $d_t^{AL}(x_t)$ by $\tau \geq 0$, which is the percentage of annotated data in the total data. Given a function $\beta(\tau)=e^{\tau-T_0}$ parameterized by a hyper-parameter $T_0 \in [0,1]$ and the total amount of dataset $|X|$, the weight of $x_t\in X$ is given by the following equation.
\begin{equation}
\small
\begin{split}
d_t^{bi}(x_t)&=\left (d_t^{AL}(x_t) \right)^{\eta(t)} * d_t^{EAT}(x_t)\\ \eta(t)&=\beta(t/|X|).
\end{split}
\end{equation}
As such, the bi-weighting mechanism highlights the importance of $d_t^{AL}(x_t)$ in the early stages of the annotation process (i.e., when $\tau < T_0$) and decreases it as the annotation process progresses. This allows for efficient use of the AL score. Notably, the bi-weighting mechanism is efficient because it allows for the fusion of AL and EAT in a post-hoc manner.

% \subsection{Learning of the model annotator}\dy{better merged into 3.4}

% We acknowledge that a more sophisticated method can be designed in future work.

\textbf{Optimization of SANT}. We optimize the model annotator $f_t$, EAT $d_t^{EAT}$, and AL method $d_t^{AL}$ independently. The loss function for AL, denoted as $L_{AL}$ (if applicable), is included in the overall loss function $\mathcal{L}$ of SANT, which is the sum of $L_{f}$, $L_{EAT}$, and $L_{AL}$. Note that all human-annotated data from past human-model interactions are utilized for optimization. For optimization details, please refer to Appendix \ref{syenx}. 

\section{Evaluation}
\label{fei78}
\begin{figure*}[!htb]
    \setlength{\abovecaptionskip}{2pt}   
    \setlength{\belowcaptionskip}{2pt}
   \centering  
   \includegraphics[width=1\textwidth]{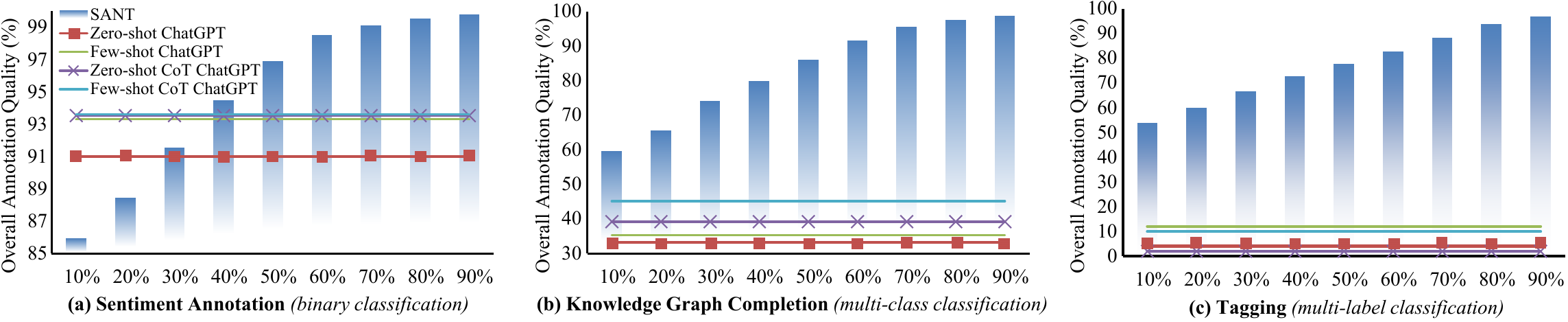} 
   \caption{Annotation quality of SANT and ChatGPT-based automatic annotation. The X-axis means the proportion of annotation budgets. As the annotation tasks become increasingly difficult (from task a to task c), human experts are indispensable to achieving high-quality annotation, despite the efficiency of adopting LLMs as annotators.}  
   \vspace{-3mm}
   \label{demogr2aph33dd3}
\end{figure*}

In this section, we conduct extensive experiments to assess the effectiveness of SANT. Given limited annotation budgets, we evaluate if SANT is more desirable to obtain high-quality annotations, compared to existing annotation methods (cf. Section \ref{dyem}). Furthermore, we comprehensively analyze the advantages of SANT and uncover the characteristics of triage-to-human and triage-to-model data in data annotation task (cf. Section \ref{dyem2}). 
% Due the space limit, the ablation on the max-margin loss $L_m$ is provided in Appendix \ref{sensi}, and practical engineering-related suggestions are in Appendix \ref{impleod}. 
% \dy{B\&F why here? F repeat to implementation details}
% CHEN: I moved them to other places

\subsection{Experiment Setup}
\label{dddd8768}
\textbf{Tasks \& Datasets}. To assess SANT, we perform three annotation tasks, namely sentiment annotation, knowledge graph completion, and tagging. These tasks have been highlighted as crucial in various web applications %papers 
% \dy{web applications}
~\cite{yao2021interactive, marcos2020information, tovstogan2020web, lee2022promptiverse}. Each task corresponds to a different type of classification problem: binary classification, multi-class classification, and multi-label classification, respectively. The difficulty level of these tasks ranges from easy to hard.
To simulate the annotation process, we utilize three commonly used datasets: IMDB for sentiment annotation, WN18RR for knowledge graph completion, and CiteULike for scientific paper tagging. More details about the datasets and tasks can be found in Appendix \ref{datainfo}.
%This implies that human does not make mistakes and we leave the error-prone human annotations for future study. 
% For the \underline{sentiment annotation}, the annotator is required to label the binary sentiment based on the input movie review. As for the \underline{knowledge graph completion}, the annotator is required to label the semantic relation of an input word pair. As for the \underline{tagging} task, the annotator is required to label the tags of a scientific paper. 

% \footnote{It is believed the model generalization is negatively correlated with the label noise rate of its training data \cite{liu2021understanding}.}
\noindent\textbf{Evaluation Metrics.} We have selected appropriate metrics to evaluate the quality of annotations. For sentiment annotation and knowledge graph completion tasks, we use accuracy as the metric. Similarly, we use Hit Ratio ($HR@10$) for the multi-label tagging task, as suggested in \citet{chen2020jit2r}. 
%Our evaluation not only focuses on the annotation quality of overall data (including expert-annotated and model-annotated data) for each annotation method but also pays special attention to the annotation quality of model-annotated data. This approach provides a more intuitive measure of the effectiveness of different methods.
% \dy{Our evaluation pays special attention to the annotation quality of model-annotated data, but also discusses the annotation quality of overall data (including expert-annotated and model-annotated data).}
Our evaluation pays special attention to the annotation quality of model-annotated data, but also discusses the annotation quality of overall data (including expert-annotated and model-annotated data).

% We value the annotation quality of the overall data and the model-annotated data. 

%To ease the optimize, We also relax $\mathbbm{1}[y_t \neq max(f_t(x_t))]$ to $\mathbbm{1}[HR@10 > 0]$ in $L_d$ during the optimization of multi-label tagging annotator.

\noindent\textbf{Comparative methods.} For a fair comparison, we use the same model annotator for both SANT and semi-automatic annotation, with the only difference being the addition of a data triage module. We consider both classic and advanced active learning methods, which focus on triage-to-human data, and \textbf{Random} strategy that allocates data randomly.
\begin{itemize}[leftmargin=*, itemsep=-4pt]
    \item \textbf{MaxEntropy} uses entropy-based strategy as data triage to assign data with high uncertainty to the expert \cite{xiao-etal-2023-freeal, li2023coannotating}.     
    \item \textbf{Calibrated MaxEntropy} aims to measure uncertainty more precisely. We equip \textit{MaxEntropy} with the model calibration, implemented by the temperature scaling \cite{guo2017calibration} with the scalar default parameter (i.e., $1.5$).
    
    \item \textbf{Ent-gn} uses an advanced AL method \cite{wang2022boosting} as the data triage module. It selects data that leads to a lower upper-bound of test loss, where the entropy is used to compute the gradient norm in an unsupervised way.
    
    \item \textbf{Exp-gn} uses labels from past interactions to compute an expected empirical loss \cite{wang2022boosting} instead of using entropy in \textit{Ent-gn}.

    \item \textbf{LLM-based automatic annotation} involves using ChatGPT as a strong baseline for automatic annotation. We test various prompts, including zero-shot, few-shot, zero-shot CoT, and few-shot CoT. Refer to Appendix \ref{chatgpt} for more details.
    
\end{itemize}
To analyze SANT's advantages and uncover its characteristics, we consider the following ablation baselines used in Section \ref{dyem2}.
\begin{itemize}[leftmargin=*, itemsep=-4pt]
    \item \textbf{SANT w/o AL} prioritizes the triage-to-model data only. It assigns easy-to-predict data to the model and hard-to-predict data to the expert.
    \item \textbf{SANT w/o EAT} prioritizes the triage-to-human data only, which is essentially semi-automatic annotation with the AL enhancement. The off-the-shelf Exp-gn is used in this case.
\end{itemize}

\noindent\textbf{Implementation details.} 
Considering the human-model annotation efficiency, light-weighted annotation models are always preferred \cite{pmlr-v133-desmond21a, chen2020jit2r, hedderich2021anea}. 
Regarding the semi-automatic annotation implementation, we use customized model annotators for different annotation tasks. Specifically, we use FastText for sentiment annotation \cite{hedderich2021anea}, the distributional model for the knowledge graph completion task \cite{roller2014inclusive, kober2021data}, and the item tagging model for tagging \cite{chen2020jit2r}. 
% When implementing AL, a dynamic mechanism \cite{wilder2021learning} is used to assign the input data, rather than adjusting the threshold. Specifically, an data $x_t$ is assigned to human if $(1-d_t^{AL}(x_t))\max(f_t(x_t)) < d_t^{AL}(x_t)$ holds. Here, $\max(\cdot)$ outputs the maximal value of the input vector. By this means, the human receives data with sufficiently high AL score, or sufficiently uncertain model output.
Regarding the implementation of SANT, MLP is used as the backbone of EAT following \citet{chen2020jit2r}. The hyper-parameter $T_0$ in the bi-weighting mechanism is set to be $0.2$ for all experiments without tuning. 
To simulate human annotation behavior, we mask out the ground truth in the datasets. the ground truth is revealed when the expert is selected by the data triage module. Moreover, to simulate the limited annotation budgets setting, we tune the annotation budgets from 10\% to 90\% with a step size being 10\%, where 10\% budgets mean the human can only annotate 10\% data. 
More implementation details are shown in Appendix \ref{thisisaim}, and engineering-related suggestions are in Appendix \ref{impleod}.

\begin{table*}[t]
\centering
\resizebox{0.98\textwidth}{!}{
\begin{tabular}{l|l|l|lllllllll}
\toprule
\textbf{Task} & \textbf{Triage Method} & \textbf{Triage Types} & 10\% & 20\% & 30\% & 40\% & 50\% & 60\% & 70\% & 80\% & 90\% \\

\midrule
\multirow{6}{*}{\begin{tabular}[c]{@{}c@{}}Sentiment\\ Annotation\\\textit{(binary} \\ \textit{classification)}\end{tabular}} 
&Random& Random &  82.90 & 82.38 & 82.24 & 82.74 & 82.15 & 82.23 & 82.27 & 82.49 & 82.44 \\
&Calibrated MaxEntropy & Classic AL  & {84.28} & {85.42} & 87.45 & 89.31 & 92.58 & {96.17} & 96.17 & 96.10 & 96.16 \\
&MaxEntropy  & Classic AL& 83.15 & 85.09 & 87.03 & 89.61 & \underline{93.21} & 95.94 & 95.93 & 95.91 & 95.93 \\
&Ent-gn  & Advanced AL&  \underline{84.37}&\textbf{85.59}&\underline{87.53}&\underline{90.76}&93.00&\textbf{96.32}&\underline{96.33}&96.32&\underline{96.47} \\
&Exp-gn  & Advanced AL&  \textbf{84.38}&\underline{85.57}&87.47&\underline{90.76}&92.98& \underline{96.29}&\underline{96.33}&\underline{96.40}&96.43 \\\cline{2-12}
&SANT (\textit{w/ Exp-gn})  & Advanced AL + EAT & \textbf{84.38}&\underline{85.57}&\textbf{87.93}&\textbf{90.79}&\textbf{93.77}&96.28&\textbf{97.00}&\textbf{97.67}&\textbf{97.77}\\
\midrule

\multirow{6}{*}{\begin{tabular}[c]{@{}c@{}}Knowledge\\Graph\\Completion\\ \textit{(multi-class} \\ \textit{classification)}\end{tabular}} 
&Random & Random&  55.97 & 55.78 & 55.92 & 55.81 & 55.87 & 57.08 & 57.03 & 57.06 & 56.82 \\
&Calibrated MaxEntropy  & Classic AL & \underline{56.66} & \textbf{58.91} & 60.75 & \underline{64.17} & 64.48 & 64.72 & 64.59 & 64.59 & 64.29 \\
&MaxEntropy  & Classic AL & \textbf{56.88} & \underline{58.90} & \underline{61.27} & 62.95 & 63.08 & 63.92 & 64.13 & 63.18 & 62.89 \\
&Ent-gn  & Advanced AL & 54.48&56.08&59.34&60.15&65.19&\underline{68.71}&75.49&78.09&79.41 \\ 
&Exp-gn  & Advanced AL & 55.14&56.81&60.28&61.51&\underline{65.40}&68.07&\underline{76.90}&\underline{81.76}&\underline{81.41} \\ \cline{2-12} 
&SANT (\textit{w/ Exp-gn})  & Advanced AL + EAT &  55.14&56.81&\textbf{62.83}&\textbf{66.32}&\textbf{71.78}&\textbf{78.67}&\textbf{85.03}&\textbf{87.31}&\textbf{87.16} \\
\midrule

\multirow{6}{*}{\begin{tabular}[c]{@{}c@{}}Tagging\\ \textit{(multi-label} \\ \textit{classification)}\end{tabular}} 
&Random & Random &  48.78 & 49.45 & 50.28 & 51.75 & \underline{49.58} & 50.48 & 49.86 & 50.11 & 55.06 \\ 
&Calibrated MaxEntropy & Classic AL  & 48.78 & 48.13 & 48.06 & 48.20 & 48.27 & 48.99 & 49.55 & 49.28 & 49.37 \\
&MaxEntropy  & Classic AL & 48.78 & 48.20 & 48.40 & 48.40 & 48.47 & 48.68 & 49.24 & 49.27 & 49.35 \\
&Ent-gn  & Advanced AL & 48.78&\underline{49.88}&\underline{51.34}&\underline{52.94}&49.52&\underline{52.02}&\underline{54.80}&\underline{56.43}&\underline{58.43} \\
&Exp-gn  & Advanced AL  &48.78&\textbf{49.90}&50.94&52.89&49.50&50.24&54.48&55.47&57.84 \\ \cline{2-12}
&SANT (\textit{w/ Exp-gn})  & Advanced AL + EAT &  48.78&\textbf{49.90}&\textbf{52.24}&\textbf{54.35}&\textbf{55.25}&\textbf{56.60}&\textbf{61.13}&\textbf{68.24}&\textbf{68.53} \\
\bottomrule
\end{tabular}
}
\caption{The annotation quality on model-annotated data (\%). The best performance is marked in \textbf{bold} and second best is \underline{underlined}. Overall, SANT enjoys higher-quality annotations. We analysis the cons of SANT on very limited data in Section \ref{dyem2}. More experiments are provided in Appendix, Table \ref{tab:moremoer}}
\label{tab:main_results4rfsd}
\vspace{-3mm}
\end{table*}

\subsection{Annotation Quality Evaluation}
\label{dyem}

%when the annotation budget is extremely limited, focusing on \textit{triage-to-human data} is more important for optimizing the utilization of the annotation budget. AL, known for its high sample-efficiency, ensures that the model can efficiently learn from extremely limited human-annotated data, leading to an overall improvement in annotation quality. 
%On the other hand, as the budget increases, the focus should be shifted to the \textit{triage-to-model data}. By carefully triaging easy-to-predict data to the model, the model annotation accuracy is significantly improved, and the overall annotation quality even exceeds that of focusing on triage-to-human data. 

This section presents a comparison of the annotation quality of SANT with LLM-based automatic methods and semi-automatic methods. The results are presented in Figure \ref{demogr2aph33dd3} and Table \ref{tab:main_results4rfsd}. % The detailed observations are provided below.
% The results, as shown in Figure \ref{demogr2aph33dd3} and Table \ref{tab:main_results4rfsd}, demonstrate that SANT consistently improves annotation quality across various annotation budgets. This highlights the importance of proper allocation of data to both human and model worker in achieving higher-quality annotation. More detailed observations can be found below.

%\textbf{Experts are indispensable to achieving high-quality annotation}. 
\textbf{Despite the efficiency of adopting LLMs as annotators, human experts are indispensable to achieving high-quality annotation.}
As illustrated in Figure \ref{demogr2aph33dd3}, while ChatGPT's performance on simple annotation tasks (i.e., sentiment annotation) is comparable to SANT, it falls behind SANT significantly when it comes to complex annotation tasks (i.e., knowledge graph completion and tagging). 
% Despite ChatGPT's strong in-context abilities, it fails to match the performance of SANT on these tasks. 
In particular, ChatGPT struggles with the tagging task due to the large number of candidate tags (i.e., 500). It often generates non-existent tags. 
% While providing few-shot examples can help alleviate this problem, accurately selecting relevant tags remains challenging.
Compared to SANT, which uses a light-weighted model architecture and a certain amount of annotation budgets, we suggest that experts play an indispensable role in achieving high-quality annotations, especially for challenging annotation tasks. Therefore, significant effort should be invested in researching methods that make full use of limited budgets to achieve high-quality annotations, which is our motivation. % Our SANT in Figure \ref{demogr2aph33dd3} achieves more than 96\% accuracy in sentiment annotation while using only 50\% of annotation budgets. Its performance on the other two tasks has more obvious advantages compared to ChatGPT.

\textbf{SANT secures higher-quality annotations compared to semi-automatic annotation}. 
Table \ref{tab:main_results4rfsd} shows the annotation quality of SANT and baselines on the model-annotated data, excluding the influence of expert annotation.
Compared to random-based method, almost all triage-based methods could achieve performance improvements. Such an exception motivates the necessity of clear division of duty between the human and model. 
%Instead of the random assignment, it is important to assign the right data to the right annotator (the expert or the model).
This is beneficial for maximizing the annotation quality and optimizing the utilization of the limited budgets.
Compared to AL-based semi-automatic annotation, SANT secures higher quality.
% On average, SANT outperforms the best AL-based method by +0.50\% for sentiment annotation, +4.86\% for knowledge graph completion, and +4.54\% for the tagging tasks across various annotation budgets.
% Comparing SANT with \textit{Exp-gn}, which shares the same model annotator and AL-based mechanism, further demonstrates the advantages of the proposed mechanisms in SANT. 
Moreover, SANT consistently outperforms \textit{Exp-gn}, which shares the same model annotator and AL-based mechanism, across different annotation budgets
% , resulting in an average performance gain of +0.51\% for sentiment annotation, +4.86\% for knowledge graph completion, and +5.00\% for the tagging tasks. 
As the annotation budgets increase, the performance gain of SANT becomes more significant due to better training of the EAT and AL components. These experiments highlight that solely improving model performance through triage-to-human data (AL) may not be sufficient for optimizing annotation quality under limited budgets. Determining which data should be assigned to the model is crucial for improving annotation quality.

\subsection{Analysis and Characteristics of SANT}
\label{dyem2}
\begin{figure*}
\centering
\setlength{\abovecaptionskip}{3pt}   
\setlength{\belowcaptionskip}{2pt}
\begin{tabular}{ccc}
\includegraphics[width=0.31\textwidth]{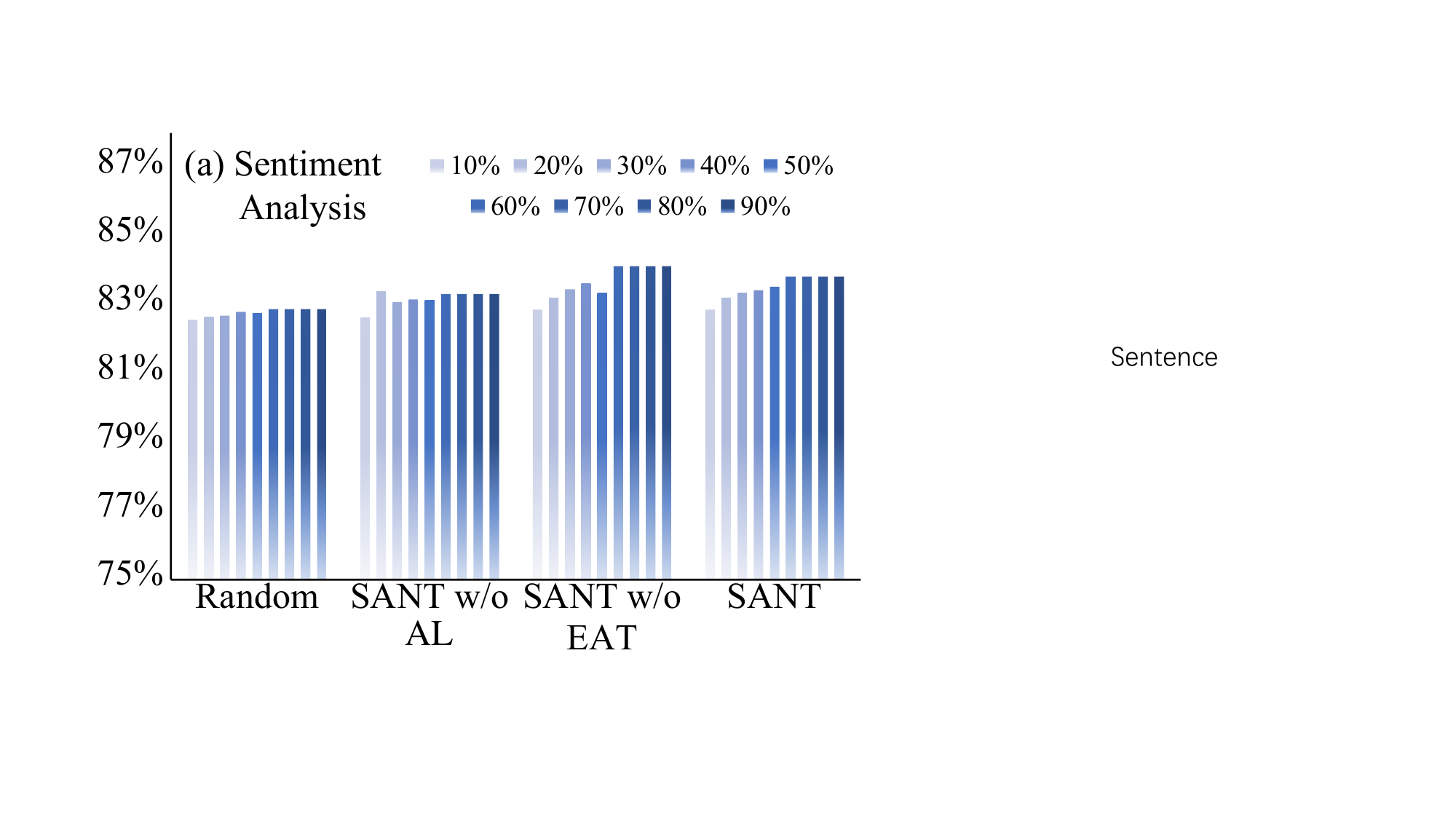}&
\includegraphics[width=0.31\textwidth]{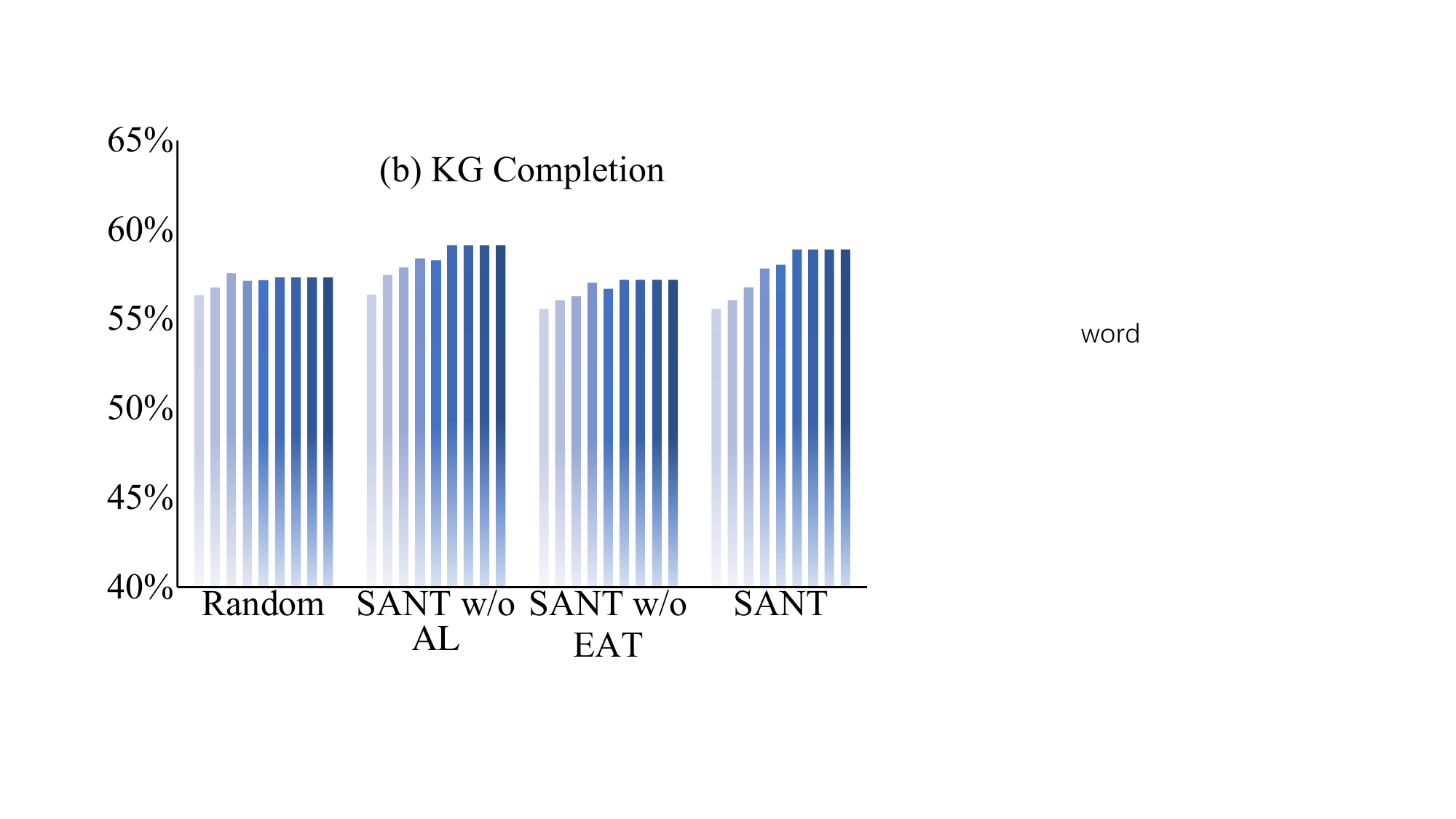}&
\includegraphics[width=0.31\textwidth]{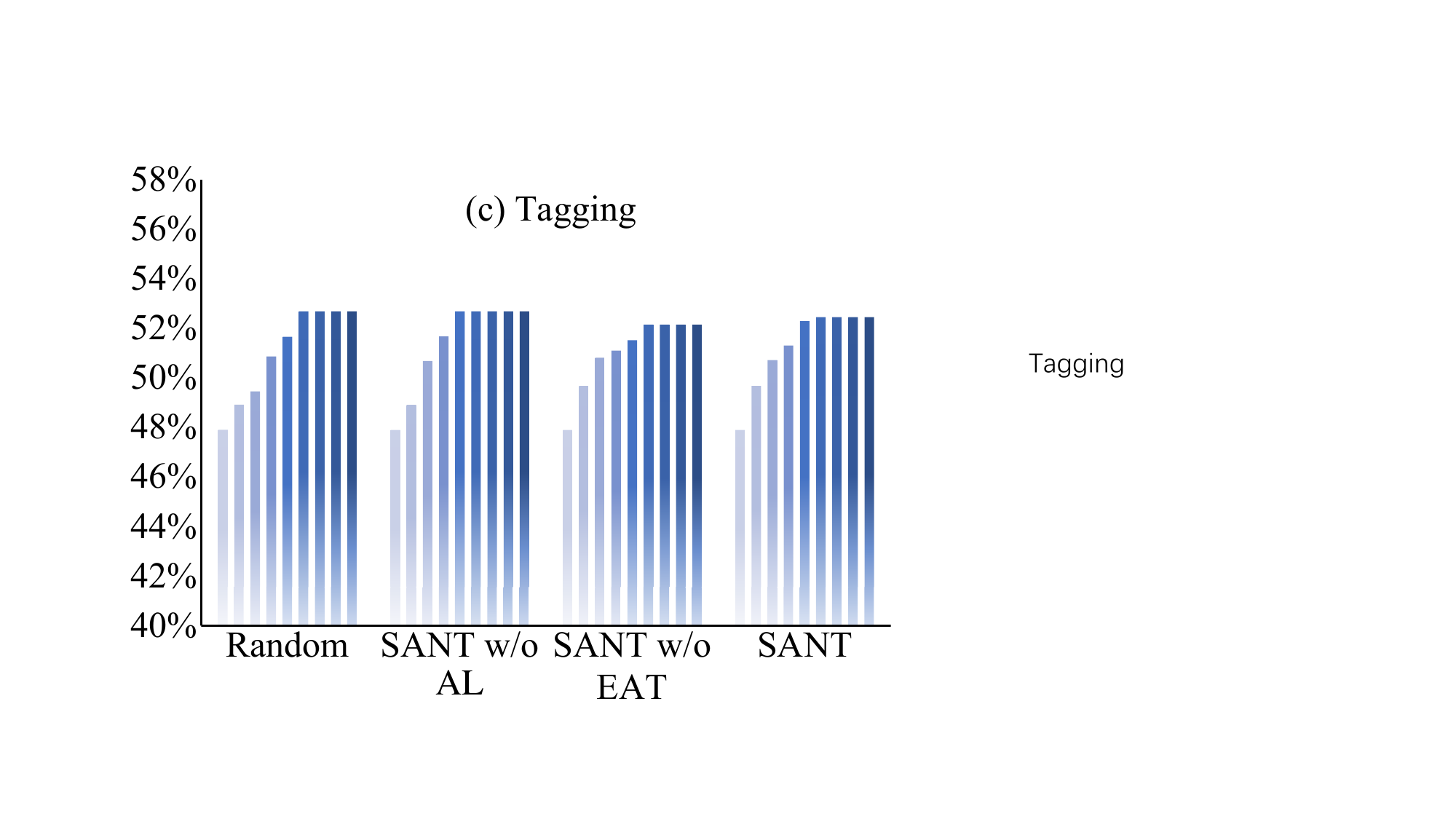}\\
% {\small (a) Sentiment annotation} & {\small (b) Knowledge graph completion} & {\small (c) Tagging}
\end{tabular}
\vspace{-1mm}
\caption {Model predictive ability evaluation on the same extra test dataset. While prioritizing triage-to-human data by AL has some advantages over triage-to-model data in promoting model predictive ability, it is not always the case (e.g., SANT w/o EAT loses its advantage in knowledge graph completion task).}
\label{fig:ideesssad4d}
\vspace{-2mm}
\end{figure*}

We aim to find the key predictor for the annotation prediction of SANT: triage-to-human data or triage-to-model data. 
The former improves the predictive ability of the model via prioritizing informative data selected by AL, while the latter carefully selects easy-to-predict data for model annotation and prioritizes hard ones for expert annotation via EAT.
We consider two ablation baselines that separate the impact of AL and EAT. 
The results are shown in Figure \ref{fig:ideesssad4d} and Table \ref{tab:main_results4rfsdq3wed}.
% The detailed observations are provided below.

\begin{table*}[t]
\centering
\resizebox{0.98\textwidth}{!}{
\begin{tabular}{l|l|l|lllllllll}
\toprule
\textbf{Task} & \textbf{Annotation Method} & \textbf{Data Triage} & 10\% & 20\% & 30\% & 40\% & 50\% & 60\% & 70\% & 80\% & 90\% \\

\midrule
\multirow{3}{*}{\begin{tabular}[c]{@{}l@{}}Sentiment\\ Annotation\end{tabular}} &SANT w/o EAT  & Advanced AL&  \textbf{84.38}&\textbf{85.57}&87.47&90.76&92.98&\textbf{96.29}&96.33&96.40&96.43 \\
&SANT w/o AL  & EAT &  \underline{83.90} & \underline{85.16} & \underline{87.89} & \underline{90.78} & \underline{93.40} & 96.13 & \underline{96.96} & \underline{97.47} & \underline{97.71} \\
&SANT  & Advanced AL + EAT & \textbf{84.38}&\textbf{85.57}&\textbf{87.93}&\textbf{90.79}&\textbf{93.77}&\underline{96.28}&\textbf{97.00}&\textbf{97.67}&\textbf{97.77}\\
\midrule

\multirow{3}{*}{\begin{tabular}[c]{@{}l@{}}Knowledge\\Graph\\Completion\end{tabular}} &SANT w/o EAT  & Advanced AL & \underline{55.14}&\underline{56.81}&60.28&61.51&65.40&68.07&76.90&81.76&81.41 \\ 
&SANT w/o AL  & EAT  & \textbf{57.06} & \textbf{59.75} & \underline{62.75} & \underline{66.08} & \underline{71.61} & \underline{78.57} & \underline{84.99} & \underline{87.28} & \underline{87.12}\\
&SANT  & Advanced AL + EAT &  \underline{55.14}&\underline{56.81}&\textbf{62.83}&\textbf{66.32}&\textbf{71.78}&\textbf{78.67}&\textbf{85.03}&\textbf{87.31}&\textbf{87.16} \\
\midrule

\multirow{3}{*}{Tagging} &SANT w/o EAT  & Advanced AL  &48.78&\textbf{49.90}&50.94&52.89&49.50&50.24&54.48&55.47&57.84 \\
&SANT w/o AL  & EAT  & 48.78 & \underline{49.81} & \underline{51.14} & \underline{52.97} & \underline{55.23} & \underline{56.53} & \underline{61.08} & \underline{68.20} & \underline{68.50} \\
&SANT  & Advanced AL + EAT &  48.78&\textbf{49.90}&\textbf{52.24}&\textbf{54.35}&\textbf{55.25}&\textbf{56.60}&\textbf{61.13}&\textbf{68.24}&\textbf{68.53} \\
\bottomrule
\end{tabular}
}
\caption{The annotation quality of SANT and its ablation counterparts on the model-annotated data (\%). Importance of triage-to-human (i.e., AL) and triage-to-model data (i.e., EAT) varies at different annotation stage. SANT effectively balances the advantages of the triage-to-human and triage-to-model data by integrating EAT and AL.}
\label{tab:main_results4rfsdq3wed}
\vspace{-3mm}
\end{table*}

\textbf{Prioritizing triage-to-human data for active learning can enhance model predictions, but this advantage is not always the case.}
We first study the effect of triage-to-human data by measuring the predictive ability. However, although conducting experiments under the same annotation budgets, models are trained on different human-annotated data and tested on different unlabeled data due to different data triage preferences of AL and EAT. This hinders the measurement of predictive ability. To address this, controlled experiments are conducted to assess the predictive ability of each model on the same extra test data. The results are illustrated in Figure \ref{fig:ideesssad4d}.
AL performs better in predicting sentiment annotations but loses its advantages in completing knowledge graph tasks. On the other hand, EAT, which implies the anti-curriculum learning \cite{braun2017curriculum}, or hard example mining \cite{shrivastava2016training}, achieves surprisingly good results in knowledge graph completion. Although prioritizing hard instances has been shown effective in training models \cite{mindermann2022prioritized, shrivastava2016training}, our experiment does not find statistically significant evidence of its effectiveness. Therefore, determining which instances should be triaged to humans for training a good model remains a challenge. Our results suggest that future studies should consider incorporating hard instances.

\textbf{Importance of triage-to-human and triage-to-model data varies at different annotation stage, which is effectively balanced by SANT.}
We further take both the triage-to-human and triage-to-model data into account simultaneously.
Table \ref{tab:main_results4rfsdq3wed} shows the annotation quality of SANT and its ablations on the model-annotated data\footnote{The results of annotation quality on the overall data are available in Appendix \ref{machine-ann}. }. 
When working with very few budgets (10\%-20\%), using AL to utilize triage-to-human data is the optimal choice as \textit{SANT w/o EAT} achieves higher quality on two-thirds of the datasets compared to \textit{SANT w/o AL}. On the contrary, when the budgets increase, EAT, which estimates model misclassification, tends to be a more stable and accurate method for human-model collaborative annotation compared to AL. This shifts the focus to triage-to-model data. In conjunction with EAT, \textit{SANT w/o AL} achieves the better results compared to \textit{SANT w/o EAT}, particularly in the middle range of annotation budgets (40\%-70\%). 
Therefore, the importance of triage-to-human and triage-to-model data varies at different annotation stage. 
To bring together two types of data, SANT utilizes the bi-weighting mechanism. We found that the annotation quality of SANT is optimal in most cases, indicating that SANT effectively allocates triage-to-human data to train the model and enhance its predictive capability, while also allocating triage-to-model data to improve the model annotation quality by providing easy-to-predict data for the model.

\textbf{The key factor influencing the quality of annotations in SANT lies in the careful selection of data to be assigned for model annotation.} 
According to Figure \ref{fig:ideesssad4d}, prioritizing triage-to-human data using AL or EAT does not lead to a significant improvement in the model's predictive ability. 
While determining which instances should be triaged to human is crucial for model training, our experiment does not find statistically significant evidence of its advantages. It is possible that the potential values of data have not been fully utilized \cite{tang2019self, mindermann2022prioritized}. Therefore, triage-to-human data is expected to have a limited contribution to improving annotation quality in SANT.
Additionally, our analysis of Table \ref{tab:main_results4rfsdq3wed} shows that SANT w/o AL has a more significant advantage in model annotation data quality compared to SANT w/o EAT. 
Therefore, we argue that carefully selecting easy-to-predict data for model prediction is crucial in significantly improving overall annotation accuracy, even though training the model on manual labeled hard-to-predict data may not result in a stronger predictive ability. This finding highlights that deciding which data to use for model prediction is more important than training itself, particularly as budgets increase. In some cases, the improvement in annotation quality can compensate for the disadvantages of an under-trained model with lower predictive ability. %We safely conclude that focusing on triage-to-model data is a more critical factor in SANT's success. 
To sum up, prioritizing triage-to-model data is a pivotal factor in determining the success of SANT.
% \vspace{-10mm}

\section{Conclusion}
We present the first work to study the roles of both triage-to-human and triage-to-model data. 
We formulate semi-automatic data annotation as a data triage problem via SANT, achieved by optimizing two novel mechanisms (i.e., EAT and bi-weighting mechanisms) in utilizing limited annotation budgets.
Empirical experiments show that SANT significantly outperforms the existing semi-automatic annotation methods.
%and revisit the role of the human and the model in the data annotation.
Although the effectiveness of LLM-based annotator, we find that experts are still indispensable to achieving high-quality annotations, especially on challenging data annotation tasks. Coordinately delegating the human and the model as a team can be a feasible solution.
We also observe that relying on triage-to-human data to create a better model may not be the most effective approach for high-quality annotations. It may be more beneficial to identify triage-to-model data. % In the future, we urge researchers to devote more attention to exploring triage-based annotation methods.%, with a particular focus on developing more effective triage strategies.

\section*{Limitations}

\textbf{Scale of Model Annotator}. One limitation of our work is that we use lightweight models for annotation due to limited computational resources. While lightweight models are commonly employed in data annotation \cite{chen2020jit2r, pmlr-v133-desmond21a}, we do not investigate the effectiveness of larger language models as the model annotator in this study. However, SANT, based on lightweight models, still outperforms the LLM-based automatic annotator.

\textbf{Model annotation cost}. While the use of models for data annotation may incur computational costs, our current research focuses solely on human annotation costs, neglecting model annotation costs. This simplification, while not entirely reflective of real-world scenarios, is justified by the significantly higher cost of human annotation compared to model annotation \cite{li2023coannotating, gilardi2023chatgpt}. Additionally, the light-weighted models employed in this study further minimize the model annotation cost, rendering it negligible. To enhance the practicality of our work, future research will incorporate the cost of model annotation.

\textbf{Human annotation cost simulation}. Human annotation cost on each piece of data may not necessarily be constant. The cognitive abilities of different humans and the difficulty level of different data can both impact the human annotation cost. Our paper, like the previous research, assumes that the human annotation cost on each data is constant. In future research, constructing a user simulator to simulate human cognitive abilities and evaluating the difficulty of the data could be the next research direction to improve our work. 

\textbf{Performance under extremely limited budgets}. Our method may not exhibit a substantial advantage over the best baseline under extremely limited budgets. However, it consistently performs at least on par, and often slightly better in this scenario. This, however, does not diminish the substantial effectiveness advantage demonstrated by our method in other scenarios. We have conducted a thorough analysis of the reasons behind our method's limited performance improvement under extremely limited budgets. This analysis will guide our future work in developing targeted improvements to further enhance the effectiveness of our method.

\textbf{Data annotation can be an intricate and complex engineering problem}. While our work focuses on the data allocation problem and introduces the SANT framework for human-machine cooperative annotation, real-world deployment of SANT necessitates further systematic engineering considerations. These aspects, often overlooked in both our paper and other research on data allocation or data annotation, are addressed in Appendix \ref{impleod}, where deployment suggestions are provided.

\section*{Acknowledgements}
This work was supported in part by the National Natural Science Foundation of China (No. 62272330); in part by the Fundamental Research Funds for the Central Universities (No. YJ202219); in part by the Singapore Ministry of Education (MOE) Academic Research Fund (AcRF) Tier 1 grant (No. MSS24C004).

\bibliography{custom}

\appendix

\section{More Related Work}
\label{morerelate}
Our research is also tied to data annotation and active learning. Furthermore, we address the data triage problem by utilizing the idea of learning to defer. To elaborate on these approaches, we provide a literature review and highlight the differences from existing methods.

\textbf{Data Annotation}. Computer-assisted annotation (or interactive annotation) becomes an alternative option. Such methods \todo{enable a model to facilitate human annotation} by providing annotation suggestions on the fly \cite{ klie-etal-2018-inception, lohr-etal-2019-continuous, marchal2021semi}. However, \todo{these methods still require manual validation on each prediction, which is inefficient and time-consuming.} Thus, it is not feasible for the setting with limited annotation budgets. 

\textbf{Active Learning}. It targets to select informative or representative data that can maximize the performance of a model \cite{Ren2020ASO}. One way to achieve this is through the uncertainty-based method \cite{desmond2021semi}, which selects data with high prediction uncertainty for human annotation. Density-based methods are also commonly used, but recent studies show that these two methods are highly correlated \cite{loquercio2020general}.
Additionally, research on selective sampling tries to combine active learning and online learning \cite{cesa2006worst, cesa2009robust}, where the model is allowed to adaptively query the label of an observed data on the fly, according to its potential to improve the model's performance. Basically, selective sampling has shown its effectiveness in online large-scale kernel \cite{wang2012breaking} and graph-based online learning \cite{9746661}. 
% The idea of using AL as the triage method is recently proposed \cite{zhang2021human}. However, they focus on the dialogue evaluation task rather than the data annotation task. 
Semi-automatic annotation utilizes the online active learning (or selective sampling) methods to triage data for expert annotations and efficiently train a good model. However, our SANT goes beyond active learning by introducing the EAT mechanism, which identifies data that the model struggles to predict. SANT then employs the bi-weighting mechanism to combine EAT and active learning, enabling it to leverage both triage-to-human data (active learning) and triage-to-model data (EAT). This integration allows SANT to take full advantage of both methods, resulting in more efficient and accurate annotations.

\textbf{Learning to Defer}. The collaboration between humans and machines can be described as "learning to defer", where tasks are jointly completed by either the machine or deferred to humans for decision-making \cite{raghu2019algorithmic, okati2021differentiable, madras2018predict}. To achieve the overall goal, a triage strategy is employed to dynamically assign tasks to different collaborators, optimizing global utilization.
Depending on the customized triage strategies, existing approaches assign data with high prediction uncertainty to humans \cite{ni2019calibration}, while out-of-distribution and noisy data to humans for decision-making \cite{wang2021classification}. Additionally, \citet{jiang2018trust} assign data with unreliable model predictions to humans to mitigate prediction bias.
Existing semi-automatic data annotation methods can also be incorporated into the \textit{learning to defer} framework, where the informative data are deferred to humans by AL methods. Recent work tries to assign data based on the uncertainty of LLM predictions \cite{li2023coannotating}, where the parameters of LLM are frozen during the whole annotation process, as is the uncertainty measurement method. However, this approach hinders the model's ability to improve its annotation results by learning from the data, relying heavily on LLM's data annotation capability for general tasks. As highlighted by our work and other researchers \cite{xiao-etal-2023-freeal}, the annotation outcomes of LLM in general tasks are not satisfying. 
% In this paper, we take the first step in re-formulating semi-automatic annotation as a data triage problem under limited budgets, solving by \textit{learning to defer}. 
To update the annotation model on the fly, we notice that applying "learning to defer" to data annotation tasks is particularly challenging. This is because existing works on learning to defer primarily focus on fully supervised settings where labels are available, whereas in the annotation task, labels are only accessible when a human is selected for annotation. This motivates us to develop a model that can learn and generalize from limited labeled data on the fly  (cf. Section \ref{7ujm0ok}). Furthermore, we propose two tailored mechanisms, i.e., EAT and bi-weighting mechanisms, and experimentally reveal some insights for the data annotation task.

\section{More Evaluation Results of SANT on Overall Dataset}
\label{machine-ann}
Tables \ref{tab:main_results4rfsd234} and \ref{tab:main_results4rfsdq3wed222} demonstrate the annotation quality of each method on the overall data. These findings align with those in Tables \ref{tab:main_results4rfsd} and \ref{tab:main_results4rfsdq3wed}, indicating that proper allocation of data to both human and model workers results in higher-quality annotation. Moreover, SANT effectively balances the advantages of triage-to-human data and triage-to-model data by allocating triage-to-model data to improve model annotation quality through easy-to-predict instances. 

Moreover, we report the more results of model annotation quality on two datasets, including SST-5 and FreeBase in Table \ref{tab:moremoer}. Due to financial limits, we compare our method with the advanced AL method (Exp-gn) and the random-based method. Based on the table results, it is clear that our method continues to be effective. Moreover, when compared to LLM-based annotators, our method can generate higher-quality annotations with some human assistance. This highlights the importance of human experts in achieving high annotation quality, even though LLMs may be efficient annotators. In the revision, we will conduct more experiments to provide stronger validation.

\begin{table*}[!htb]
\centering
\resizebox{0.98\textwidth}{!}{
\begin{tabular}{l|l|l|lllllllll}
\toprule
\textbf{Task} & \textbf{Annotation Method} & \textbf{Data Triage} & 10\% & 20\% & 30\% & 40\% & 50\% & 60\% & 70\% & 80\% & 90\% \\

\midrule
\multirow{6}{*}{\begin{tabular}[c]{@{}c@{}}Sentiment\\ Annotation\\\textit{(binary} \\ \textit{classification)}\end{tabular}} 
&Random& Random & 84.61 & 85.90 & 87.57 & 89.64 & 91.08 & 92.89 & 94.68 & 96.50 & 98.24 \\
&Calibrated MaxEntropy & Classic AL  & 85.85 & 88.34 & 91.22 & 93.59 & 96.29 & 98.47 & 98.85 & 99.22 & 99.62 \\
&MaxEntropy & Classic AL& 84.84 & 88.07 & 90.92 & 93.77 & \underline{96.61} & 98.38 & 98.78 & 99.18 & 99.59 \\
&Ent-gn & Advanced AL& \underline{85.93} & \textbf{88.47} & \underline{91.27} & \underline{94.46} & 96.50 & 98.53 & \underline{98.90} & 99.26 & \underline{99.65} \\
&Exp-gn & Advanced AL& \textbf{85.94} & \underline{88.46} & 91.23 & \underline{94.46} & 96.49 & \textbf{98.52} & \underline{98.90} & \underline{99.28} & 99.64 \\\cline{2-12}
&SANT & Advanced AL + EAT & \textbf{85.94} & \underline{88.46} & \textbf{91.55} & \textbf{94.47} & \textbf{96.89} & \underline{98.51} & \textbf{99.10} & \textbf{99.53} & \textbf{99.78} \\ 
\midrule

\multirow{6}{*}{\begin{tabular}[c]{@{}c@{}}Knowledge\\Graph\\Completion\\ \textit{(multi-class} \\ \textit{classification)}\end{tabular}} 
&Random & Random&60.37 & {64.62} & {69.14} & {73.49} & {77.94} & {82.83} & {87.11} & {91.41} & {95.68} \\
&Calibrated MaxEntropy & Classic AL & \underline{60.99} & \textbf{67.13} & 72.53 & \underline{78.50} & 82.24 & 85.89 & 89.38 & 92.92 & 96.43 \\
&MaxEntropy & Classic AL & \textbf{61.19} & \underline{67.12} & \underline{72.89} & 77.77 & 81.54 & 85.57 & 89.24 & 92.64 & 96.29 \\
&Ent-gn & Advanced AL & 59.03 & 64.86 & 71.54 & 76.09 & 82.60 & \underline{87.48} & 92.65 & 95.62 & 97.94 \\
&Exp-gn & Advanced AL & 59.63 & 65.45 & 72.20 & 76.91 & \underline{82.70} & 87.23 & \underline{93.07} & \underline{96.35} & \underline{98.14} \\ \cline{2-12} 
&SANT & Advanced AL + EAT & 59.63 & 65.45 & \textbf{73.98} & \textbf{79.79} & \textbf{85.89} & \textbf{91.47} & \textbf{95.51} & \textbf{97.46} & \textbf{98.72} \\

\midrule

\multirow{6}{*}{\begin{tabular}[c]{@{}c@{}}Tagging\\ \textit{(multi-label} \\ \textit{classification)}\end{tabular}} 
&Random & Random & 53.90 & 59.56 & 65.20 & 71.05 & \underline{74.79} & 80.19 & 84.96 & 90.02 & 95.51 \\
&Calibrated MaxEntropy & Classic AL & 53.90 & 58.50 & 63.64 & 68.92 & 74.14 & 79.60 & 84.87 & 89.86 & 94.94 \\
&MaxEntropy & Classic AL & 53.90 & 58.56 & 63.88 & 69.04 & 74.24 & 79.47 & 84.77 & 89.85 & 94.94 \\
&Ent-gn & Advanced AL & 53.90 & \underline{59.90} & \underline{65.94} & \underline{71.76} & 74.76 & \underline{80.81} & \underline{86.44} & \underline{91.29} & \underline{95.84} \\
&Exp-gn & Advanced AL & 53.90 & \textbf{59.92} & 65.66 & 71.73 & 74.75 & 80.10 & 86.34 & 91.09 & 95.78 \\ \cline{2-12}
&SANT & Advanced AL + EAT & 53.90 & \textbf{59.92} & \textbf{66.57} & \textbf{72.61} & \textbf{77.63} & \textbf{82.64} & \textbf{88.34} & \textbf{93.65} & \textbf{96.85} \\
\bottomrule
\end{tabular}
}
\caption{The annotation quality of SANT and baselines on overall data (\%). We consider three annotation tasks with different difficulties. The best performance is marked in \textbf{bold} and second best is underlined.}
\label{tab:main_results4rfsd234}
\end{table*}

\begin{table*}[!htb]

\centering
\resizebox{0.98\textwidth}{!}{
\begin{tabular}{c|l|l|lllllllll}
\toprule
\textbf{Task} & \textbf{Annotation Method} & \textbf{Data Triage} & 10\% & 20\% & 30\% & 40\% & 50\% & 60\% & 70\% & 80\% & 90\% \\

\midrule
\multirow{3}{*}{\begin{tabular}[c]{@{}c@{}}Sentiment\\ Annotation\end{tabular}} &SANT w/o EAT  & Advanced AL& \textbf{85.94} & \textbf{88.46} & 91.23 & \underline{94.46} & 96.49 & \textbf{98.52} & {98.90} & {99.28} & 99.64 \\
&SANT w/o AL  & EAT & \underline{85.51}&\underline{88.13} &\underline{91.52} &\textbf{94.47}&\underline{96.70}&98.45&\underline{99.09}&\underline{99.49}&\underline{99.77}   \\
&SANT  & Advanced AL + EAT & \textbf{85.94} & \textbf{88.46} & \textbf{91.55} & \textbf{94.47} & \textbf{96.89} & \underline{98.51} & \textbf{99.10} & \textbf{99.53} & \textbf{99.78} \\ 
\midrule

\multirow{3}{*}{\begin{tabular}[c]{@{}c@{}}Knowledge\\Graph\\Completion\end{tabular}} &SANT w/o EAT  & Advanced AL & \underline{59.63} & \underline{65.45} & 72.20 & 76.91 & {82.70} & 87.23 & {93.07} & \underline{96.35} & {98.14} \\ 
&SANT w/o AL  & EAT  & \textbf{61.35}&\textbf{67.80}&\underline{73.93}&\underline{79.65}&\underline{85.81}&\underline{91.43}&\underline{95.50}&\textbf{97.46}&\underline{98.71} \\
&SANT  & Advanced AL + EAT & \underline{59.63} & \underline{65.45} & \textbf{73.98} & \textbf{79.79} & \textbf{85.89} & \textbf{91.47} & \textbf{95.51} & \textbf{97.46} & \textbf{98.72} \\
\midrule

\multirow{3}{*}{Tagging} &SANT w/o EAT  & Advanced AL  & 53.90 & \textbf{59.92} & 65.66 & 71.73 & 74.75 & 80.10 & 86.34 & 91.09 & \underline{95.78} \\
&SANT w/o AL  & EAT  &53.90&\underline{59.85} &\underline{65.80} &\underline{71.78} &\underline{77.62}&\underline{82.61}&\underline{88.32}&\underline{93.64}&\textbf{96.85} \\
&SANT  & Advanced AL + EAT & 53.90 & \textbf{59.92} & \textbf{66.57} & \textbf{72.61} & \textbf{77.63} & \textbf{82.64} & \textbf{88.34} & \textbf{93.65} & \textbf{96.85} \\
\bottomrule
\end{tabular}
}
\caption{The annotation quality of SANT and its ablation counterparts on the overall data (\%).}
\label{tab:main_results4rfsdq3wed222}
\end{table*}

% \usepackage{multirow}
% \usepackage{graphicx}
\begin{table*}[]
\centering
\resizebox{\textwidth}{!}{%
\begin{tabular}{c|l|l|l|l|l|l|l|l|l|l|l}
\toprule
\multicolumn{1}{c|}{\textbf{Task}} & \textbf{Annotation Method} & \textbf{Data Triage} & \textbf{10\%} & \textbf{20\%} & \textbf{30\%} & \textbf{40\%} & \textbf{50\%} & \textbf{60\%} & \textbf{70\%} & \textbf{80\%} & \textbf{90\%} \\ \midrule
\multirow{5}{*}{\begin{tabular}[c]{@{}c@{}}\textbf{Sentiment} \\ \textbf{Annotation} \\ (\textit{multi-class} \\ \textit{annotation})\\ SST-5\end{tabular}} & Zero-shot ChatGPT & None & \multicolumn{9}{c}{41.30} \\ \cline{2-12}
 & Few-shot ChatGPT & None & \multicolumn{9}{c}{53.85} \\ \cline{2-12}
 & Random & Random & 30.02 & 33.59 & 36.64 & 37.56 & 38.23 & 38.79 & 39.33 & 39.54 & 39.95 \\ \cline{2-12}
 & Exp-gn & Advanced AL & 30.13 & 34.04 & 38.43 & 40.09 & 45.74 & 49.63 & 54.08 & 60.31 & 64.95 \\ 
 & SANT & Advanced AL + EAT & \textbf{30.13} & \textbf{34.04} & \textbf{40.77} & \textbf{43.60} & \textbf{49.20} & \textbf{54.87} & \textbf{59.81} & \textbf{65.26} & \textbf{70.51} \\ \midrule
\multirow{5}{*}{\begin{tabular}[c]{@{}c@{}}\textbf{Knowledge} \textbf{graph} \\\textbf{completion}\\ (\textit{multi-class} \\ \textit{annotation})\\ FreeBase\end{tabular}} & Zero-shot ChatGPT & None & \multicolumn{9}{c}{38.85} \\ \cline{2-12}
 & Few-shot ChatGPT & None & \multicolumn{9}{c}{51.60} \\ \cline{2-12}
 & Random & Random & 22.46 & 35.19 & 42.77 & 48.54 & 52.50 & 55.31 & 57.75 & 60.26 & 61.88 \\ 
 & Exp-gn & Advanced AL & 22.86 & 36.56 & 46.84 & 50.56 & 64.18 & 69.99 & 73.98 & 79.22 & 83.48 \\ \cline{2-12}
 & SANT (Ours) & Advanced AL + EAT & \textbf{22.86} & \textbf{36.56} & \textbf{45.03} & \textbf{52.17} & \textbf{58.79} & \textbf{74.53} & \textbf{77.04} & \textbf{83.77} & \textbf{88.13} \\ \bottomrule
\end{tabular}%
}
\caption{More results on the annotation quality of SANT and baselines on model-annotated data. Two more datasets are considered here, including the SST-5 and FreeBase. Due to financial constraint, we compare our method with the ChatGPT, advanced AL method (Exp-gn), and random-based method.}
\label{tab:moremoer}
\end{table*}

\section{Ablation study on the proposed max-margin loss $L_m$}
\label{sensi}
\begin{figure}[!htb]
\centering
\begin{tabular}{c}
\includegraphics[width=0.45\textwidth]{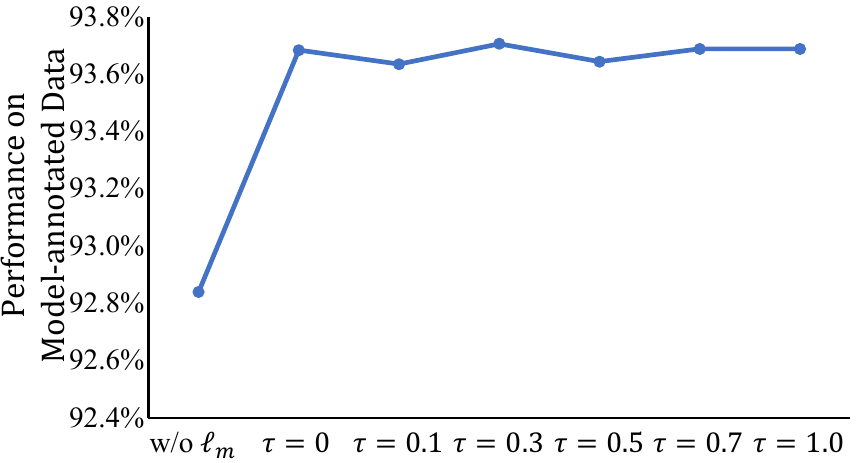}\\
\end{tabular}
\caption {Ablation study on $L_m$ and hyper-parameter tuning analysis on $\tau$. We simply choose $\tau=0.3$ in our experiments. }
\label{labors443ddddf}
%% \vspace{-5mm}
\end{figure}

As shown in Section \ref{asduionweasdo}, EAT uses the max-margin loss $L_m$ to take advantage of the partially labeled data. The loss is parameterized by a margin hyper-parameter $\tau$ to force the average loss on the assign-to-human data to be much higher than the assign-to-model data. In this section, we offer a detailed discussion of the loss itself and the sensitivity analysis. Specifically, the ablation study on $L_m$ and the hyper-parameter sensitivity study on $\tau$ are given in Figure \ref{labors443ddddf}. 
Here, we take the sentiment annotation task as an example, and require \textit{SANT w/o AL} to reject 50\% of the data to the model in each batch, making sure our ablation studies are performed on the training data of the same size. We focus on the model performances on model-annotated data, marked as $MA_{Accept}$. Here, $\tau$ is tuned in $\{0, 0.1, 0.3, 0.5, 0.7, 1.0\}$. According to the results, we believe that the max-margin loss $L_m$ helps train a better EAT. Moreover, the $L_m$ is relatively less sensitive to its hyper-parameter $\tau$. Thus, we simply choose $\tau=0.3$.

\section{Details on Datasets}
\label{datainfo}
To comprehensively evaluate SANT, we use IMDB\footnote{https://www.kaggle.com/datasets/lakshmi25npathi/imdb-dataset-of-50k-movie-reviews}, WN18RR\footnote{https://paperswithcode.com/dataset/wn18rr}, and CiteULike\footnote{Following \cite{chen2020jit2r, wang2013collaborative}, this data is used for tagging experiment. See https://github.com/js05212/citeulike-a} to simulate sentiment annotation, knowledge graph completion, and tagging tasks, respectively. Those three datasets are all English datasets and they formulate three different classification problems, including binary-class classification, multi-class classification, and multi-label classification problems. For the \underline{sentiment annotation}, the annotator is required to label the binary sentiment based on the input movie review. As for the \underline{knowledge graph completion}, the annotator is required to label the semantic relation of an input word pair. As for the \underline{tagging} task, the annotator is required to label the tags of a scientific paper.

Specifically, IMDB is a common-used data with 50K movie reviews and binary sentiment classes. WN18RR we use is a subset of WN18 with no inverse relations. WN18RR contains 93003 triples, 40943 entities, and 11 relation types. Lastly, CiteULike \cite{chen2020jit2r, wang2013collaborative} is a commonly used tagging dataset, collected from CiteULike and Google Scholar. It includes 7288 authors’ 160272 citations on 8212 papers, together with 46391 different tags. Each paper in this dataset has its abstract, title, and multiple tags. The concatenation of them are used as the input in our experiments. Following \cite{chen2020jit2r}, we also use 500 most frequent tags in our tagging experiment. 

Considering the human-model interaction efficiency, light-weighted annotation models or systems are always preferred \cite{pmlr-v133-desmond21a, chen2020jit2r}. In our experiments, we use the pre-trained GloVe\footnote{Our knowledge graph completion task is context-free. Considering the consistency of the experimental setup, we do not use contextual embedding like BERT embedding.} \cite{pennington2014glove} for all tasks to encode the word representation without fine-tuning. 
For all the dataset, we first filter out words that are not contained in the pre-trained Glove embedding. For word pairs in WN18RR, we only retain those pairs that both words have Glove embeddings. Following previous works \cite{roller2014inclusive, kober2021data}, the word pair embedding is simply obtained by concatenating the embedding of two words. As for the sentence embedding used in CiteULike and IMDB, we follow the FastText \cite{joulin2017bag} and obtain the sentence embedding by averaging the word embedding. We also remove the data if its 90\% words (or more) have no Glove embeddings.

\section{Implementation details}
\label{thisisaim}
\subsection{Implementation of SANT}
SANT is flexible to integrate fancy models. It shares the same model annotator with semi-automatic annotation. However, considering the efficiency requirement of the human-model interaction, light-weighted annotation models or systems are always preferred \cite{pmlr-v133-desmond21a, chen2020jit2r}. Referring to \citet{chen2020jit2r}, MLP is used as the EAT. For Implementation, our EAT outputs a two-dimensional predictive distribution as the triage signal rather than a real number between 0 and 1. By this means, we avoid setting a threshold to map the number to binary values of 0 and 1.
We also use the class weight to $L_d$ to handle the potential class imbalance problem and penalize mistakes in the minority class. Specifically, the weight is calculated based on the cumulative class distribution of human annotations. As for the model annotator in SANT, we use the corresponding model used in the semi-automatic annotation method. All our codes are implemented in Pytorch with one Nvidia GeForce RTX 3090. the three-layer FC with ReLu activation and dropout is utilized. Moreover, we set $k=3$ for retrieving top-k local information to integrate the input correlations. To ease the optimization, We also relax $\mathbbm{1}[y_t \neq max(f_t(x_t))]$ to $\mathbbm{1}[HR@10 > 0]$ in $L_d$ during the optimization of multi-label tagging annotator. Regarding the hyper-parameter $T_0$ in the bi-weighting mechanism, we set it to be $0.2$ for all experiments without tuning. 

Finally, to simulate the limited annotation budgets setting, we tune the annotation budgets from 10\% to 90\% with a step size of 10\%, where 10\% budgets mean the human can only annotate 10\% data. We also require the human to annotate the first 1k data to initialize the model annotator for each annotation method. Note that when it comes to the Tagging task, the first 1k data and the 10\% annotation budgets are almost equal. Therefore, in the case of 10\% annotation budgets, all the budgets (i.e., the first 1k data) are used for human annotations. The corresponding results of the different triage-based methods in Table \ref{tab:main_results4rfsd} and Table \ref{tab:main_results4rfsdq3wed} are the same.

\subsection{Implementation of Baselines}
Following \citet{wang2022boosting}, our AL-based methods update the task model on the fly and output the model at the end of interactive data selection/annotation. As for \textit{MaxEntropy} and \textit{Calibrated MaxEntropy}, we measure uncertainty in sentiment annotation and knowledge graph completion, entropy is used, following a recent annotation method \cite{desmond2021semi, li2023coannotating}. In the case of the tagging task, which is a multi-label classification task, \textit{total entropy} \cite{depeweg2018decomposition} is used.
Next, a dynamic mechanism \cite{wilder2021learning} is used to assign the input data, rather than adjusting the threshold. Specifically, an data $x_t$ is assigned to human if $(1-d_t^{AL}(x_t))\max(f_t(x_t)) < d_t^{AL}(x_t)$ holds. Here, $\max(\cdot)$ outputs the maximal value of the input vector. By this means, the human receives data with sufficiently high AL score, or sufficiently uncertain model output. As for \textit{Exp-gn} and \textit{Ent-gn}, the outputs of the active learning method \cite{wang2022boosting} can not be easily normalized. In this case, half the data in a batch with the highest AL scores\footnote{Authors show the larger the amount of data queries, the better the AL effect. Considering the human annotation labors, we assign half data to be labeled by the human in experiments.} are assigned to the human and the rest to the model. 

\subsection{Implementation of Human Simulation}
We consider human-machine cooperative data annotation under limited budgets, specifically a budget that restricts the number of human annotators that can be employed to annotate the entire dataset. In this case, our experiments focus solely on estimating the cost of human annotation, as it significantly outweighs the cost of model annotation, as evidenced by previous studies \cite{li2023coannotating, gilardi2023chatgpt}. In particular, following previous works on data annotation \cite{chen2020jit2r, pmlr-v133-desmond21a, hedderich2021anea} and human-model interaction \cite{hwa2000sample, kristjansson2004interactive}, we assume a consistent cost for human annotation per data point. This allows us to estimate the overall annotation budget using the number of human-annotated data points.
To simulate human annotation behavior, we mask out the ground truth in the datasets. the ground truth is revealed when the human is selected by the data triage module. Moreover, to simulate the constraints of limited annotation budgets, we adjust the budget allocation from 10\% to 90\% of the dataset, with increments of 10\%. This means that under a 10\% budget, human annotators can only label 10\% of the total dataset. This range of budget constraints allows us to assess the performance of SANT under varying budget limitations.

\begin{table*}
\centering
\resizebox{0.95\textwidth}{!}{
\begin{tabular}{l|ccc|ccc|ccc}
\toprule
\multirow{3}{*}{Res.}  &  \multicolumn{3}{c|}{KGC}  &  \multicolumn{3}{c|}{SA}  &  \multicolumn{3}{c}{Tagging}  \\ \cline{2-10} 
  &\multicolumn{1}{l}{\begin{tabular}[c]{@{}c@{}}SANT \\ (re-anno.)\end{tabular}}  &\multicolumn{1}{l}{\begin{tabular}[c]{@{}c@{}}SANT \\ (re-train+re-anno.)\end{tabular}}  &  SANT  &\multicolumn{1}{l}{\begin{tabular}[c]{@{}c@{}}SANT \\ (re-anno.)\end{tabular}}&\multicolumn{1}{l}{\begin{tabular}[c]{@{}c@{}}SANT \\ (re-train+re-anno.)\end{tabular}}&  SANT  &\multicolumn{1}{l}{\begin{tabular}[c]{@{}c@{}}SANT \\ (re-anno.)\end{tabular}}&\multicolumn{1}{l}{\begin{tabular}[c]{@{}c@{}}SANT \\ (re-train+re-anno.)\end{tabular}}&  SANT  \\ \midrule
10\%  &  \textbf{61.93}  &-&  59.63  &  \textbf{86.05}  &-&  85.94  &  53.90  &-&  53.90\\
20\%  &  \textbf{68.15}  &-&  65.45  &  \textbf{88.73}  &-&  88.46  &  \textbf{60.15}  &-&  59.92\\
30\%  &  \textbf{74.67}  &-&  73.98  &  \textbf{92.09}  &-&  91.55  &  \textbf{66.32}  &-&  66.57\\
40\%  &  79.72           &-&  \textbf{79.79}  &  \textbf{95.01}  &-&  94.47  &  \textbf{71.91}  &-&  72.61\\
50\%  &  \textbf{86.04}  &-&  85.89  &  \textbf{97.32}  &-&  96.89  &  \textbf{78.10}  &-&  77.63\\
60\%  &  {91.49}  &\textbf{93.93}&  91.47  &  {98.64}  &\textbf{99.09}&  98.51  &  {82.63}  &\textbf{87.20}&  82.64\\
70\%  &  {95.60}  &\textbf{95.61}&  95.51  &  {99.15}  &\textbf{99.24}&  99.10  &  {88.33}  &\textbf{89.31}&  88.34\\
80\%  &  97.46  &\textbf{97.48}&  97.46  & \textbf{99.55}  &99.53&  99.53  &  93.64  &\textbf{93.66}&  93.65\\
90\%  &  98.71  &\textbf{98.72}&  98.72  &  \textbf{99.80}  &99.76&  99.78  &  96.85  &\textbf{96.91}&  96.85\\
\bottomrule
\end{tabular}
}
\caption{The overall annotation quality of SANT after the model annotator (re-train and) re-annotating the data that are previously model-annotated. Here, '-' means no human budgets left when data run out. }
\label{tab:ydidlast}
% \vspace{-3mm}
\end{table*}

\section{Optimization Details of SANT}
\label{syenx}
In SANT, whenever the human reveals the ground truth of $x_t$, the model $f_t$ is updated with $L_f$ serving as the learning loss for the model (e.g., NLLLoss). However, our research focus is not on building a better model annotator. Therefore, we use different task-specified off-the-shelf model annotators in our experiments.

To simplify the optimization process, we optimize the model annotator $f_t$, EAT $d_t^{EAT}$, and AL method $d_t^{AL}$ independently. The loss function for AL, denoted as $L_{AL}$ (if applicable), is included in the overall loss function $\mathcal{L}$ of SANT, which is the sum of $L_{f}$, $L_{EAT}$, and $L_{AL}$. It is important to note that all human-annotated data from past human-model interactions are utilized for optimization.

During optimization, three things are taken into consideration. First, the output of the data triage module is relaxed to continuous values over $[0,1]$ to avoid the integer optimization problem. Second, the top-k retrieval and data triage modules are implemented using the Gumbel-softmax re-parameterization trick \cite{jang2016categorical} to ease the non-differentiable problem. Finally, the overall loss function is a bi-level optimization problem \cite{bard2013practical}, where optimizing the data allocation (i.e., $d_t^{AL}$ and $d_t^{EAT}$) is nested within another problem of optimizing the model annotator $f_t$. Thus, we update $f_t$ and data allocation independently and iteratively in a coordinate-descent style. Specifically, let $\theta_f^k$, $\theta_{d^{EAT}}^k$, and $\theta_{d^{AL}}^k$ be the network parameters for $f$, $d^{EAT}$, and $d^{AL}$ at optimization iteration $k$. The update procedure is given in Eq.(\ref{ddd9876d}). In the $k^{th}$ optimization iteration, we first optimize the parameter of model $f$ with the parameter of $d^{EAT}$ and $d^{AL}$ fixed. In the $(k+1)^{th}$ epoch, we exchange and optimize $d^{EAT}$ and $d^{AL}$ with $f$ fixed, and so on.
\begin{equation}
\label{ddd9876d}
\small
\begin{split}
    \theta_f^{k+1} &= \theta_f^k - \bigtriangledown_f \mathcal{L} \left (f^k, (d^{EAT})^k, (d^{AL})^k \right ),\\
    \theta_{d^{EAT}}^{k+1} &= \theta_{d^{EAT}}^k - \bigtriangledown_{d^{EAT}} \mathcal{L}\left (f^{k+1}, (d^{EAT})^k, (d^{AL})^k \right ),\\
    \theta_{d^{AL}}^{k+1} &= \theta_{d^{AL}}^k - \bigtriangledown_{d^{AL}} \mathcal{L}\left (f^{k+1}, (d^{EAT})^k, (d^{AL})^k \right ).
\end{split}
\end{equation}
In our approach, EAT is optimized towards measuring the predictive ability of $f_{t+1}$ instead of $f_t$. It is important to note that if the AL method does not contain a learnable parameter, such as entropy-based AL, it can be omitted from the optimization process.

\section{Automatic annotation via ChatGPT}
\label{chatgpt}
Since ChatGPT has demonstrated strong zero-shot, few-shot, and CoT capabilities in various domains and tasks \cite{liu2023comprehensive, hu2023zero, shah2023zero}, we use ChatGPT for automatic data annotation experiments. Specifically, we test various prompting methods for ChatGPT, including zero-shot, few-shot, zero-shot CoT, and few-shot CoT prompts. The prompts for three annotation tasks are provided in Table \ref{tab:my-chatgpt}. Due to the cost of using ChatGPT, we sample a subset for each task and then deliver it to ChatGPT for data annotation. Specifically, for the sentiment annotation task, we randomly sample 1000 data for ChatGPT annotation. As for the knowledge graph completion task, the WN18RR data contains multiple relation types. As a result, we choose 100 data at random from each class for ChatGPT annotation. Last, ChatGPT fails at the tagging task, because the dataset CiteULike contains 500 different tags in our experiments. Despite our several attempts, ChatGPT is unable to select one or more tags from the 500 available ones based on the semantics of the input words. Therefore, we cut down the candidates to 200. 

In practice, as ChatGPT has been banned in some countries, we use Monica (refer to \url{https://monica.im/}) for annotations, which is a Chrome extension powered by ChatGPT API. As free users of Monica, we have a daily usage limit. Thus, we require Monica to annotate multiple instances at a time (precisely, 20 at a time).

\section{Practical suggestions for the interactive annotation system implementation}
\label{impleod}
To implement SANT into an annotation system in the practical scenario, some tricks might be helpful to further reduce human annotation labor and improve the annotation quality of both the human and model. 

\subsection{Improving user experience}
Considering generating an annotated dataset is usually time-consuming and expensive for the human, building an annotation tool with a user-friendly UI into the system plays an important role. A previous study on the tool design shows that if offering a better user experience during the annotation \cite{cerezo-etal-2021-tools}, the annotation quality improves.

\subsection{Providing annotation suggestions to human}
When the human is selected to annotate data, the annotation system could attach the model predictions as suggestions to the human \cite{desmond2022ai, klie-etal-2018-inception}, avoiding the need to annotate data from scratch and thus reducing human cognitive labor.

\subsection{On error-prone human annotations}
Human annotations could be error-prone and inconsistent. It would usually hamper the overall annotation quality. Some works from other research domains are devoted to examining the error-prone human annotations, such as the crowd-sourced data annotation \cite{shirani-etal-2019-learning, larson2020inconsistencies}. Considering it is not the research focus of this paper, we suggest the reader to those works before implementing a real human-model interactive system.

\subsection{Post hoc correction for model annotation}
\label{Thdhmeuc8en}
The system might also involve more strategies to improve model annotations in a post hoc way. The system could use the model annotator from the final round, and re-annotate data, that was originally model-annotated at the early annotation rounds. If the final model in SANT is properly trained, the quality of the annotations might be further enhanced. If there are some budgets left for the human to annotate, the final model could be re-trained before the re-annotating, together with the newly human-annotated data. 

Taking SANT for example, Table \ref{tab:ydidlast} demonstrates the improvement in annotation quality after the model annotator re-annotated the data that were previously model-annotated. The re-annotation does help to improve the quality, and such improvement could be further enhanced by the re-training procedure. However, model re-training and model re-annotating require extra time cost. In the real-world scenario, one should decide whether to use the two extra modules (i.e., the model re-training and model re-annotating) based on the product form of his or her annotation system.

% \usepackage{graphicx}
\begin{table*}[]
\centering
\resizebox{0.98\textwidth}{!}{%
\begin{tabular}{l|ll}
\toprule
\textbf{Task} & \textbf{Zero-shot} & \textbf{Zero-shot CoT} \\ \midrule
SA & \begin{tabular}[c]{@{}l@{}}\#\#\#\\ Read the sentences and identify their sentiment. \\ There are two options: negative or positive\\ \#\#\#\\ 1. {[}INPUT\_SENTENCE1{]}\\ 2. {[}INPUT\_SENTENCE2{]}\end{tabular} & \begin{tabular}[c]{@{}l@{}}\#\#\#\\ Read the sentences and identify their sentiment. \\ You should think step by step and output the sentiment of each sentence.\\There are two optional sentiments: negative or positive\\ \#\#\#\\ 1. {[}INPUT\_SENTENCE1{]}\\ 2. {[}INPUT\_SENTENCE2{]}\end{tabular} \\ \midrule
KGC & \begin{tabular}[c]{@{}l@{}}\#\#\#\\ Read the word pairs and identify their semantic relation. \\ There are 11 options: entailment, hyponym, hypernym\\ , member, synonym, antonym, ...\\ \#\#\#\\ 1. {[}INPUT\_WORD1{]}, {[}INPUT\_WORD2{]}\\ 2. {[}INPUT\_WORD3{]}, {[}INPUT\_WORD4{]}\end{tabular} & \begin{tabular}[c]{@{}l@{}}\#\#\#\\ Read the word pairs and identify their semantic relation. \\ You should think step by step and output the semantic relation of each sentence.\\ You have the following options: entailment, hyponym, hypernym, member, synonym, antonym, ...\\ \#\#\#\\ 1. {[}INPUT\_WORD1{]}, {[}INPUT\_WORD2{]}\\ 2. {[}INPUT\_WORD3{]}, {[}INPUT\_WORD4{]}\end{tabular} \\ \midrule
Tagging & \begin{tabular}[c]{@{}l@{}}\#\#\#\\ Read the following sentences, \\ choose one or more tags from the TagSet \\ that correspond to each sentence's semantics.\\ TagSet=\{...\}\\ \#\#\#\\ 1. {[}INPUT\_SENTENCE1{]}\\ 2. {[}INPUT\_SENTENCE2{]}\end{tabular} & \begin{tabular}[c]{@{}l@{}}\#\#\#\\ Read the following sentences, choose one or more tags from the TagSet \\ that correspond to each sentence's semantics.\\ You should think step by step and output the tag list of each sentence.\\ TagSet=\{...\}\\ \#\#\#\\ 1. {[}INPUT\_SENTENCE1{]}\\ 2. {[}INPUT\_SENTENCE2{]}\end{tabular} \\ \bottomrule

\textbf{Task} & \textbf{Few-shot} & \textbf{Few-shot CoT} \\ \midrule          
SA & \begin{tabular}[c]{@{}l@{}}\#\#\#\\ Read the sentences and identify their sentiment. \\ There are two options: negative or positive\\  Here are two examples that you can refer to.\\ Example 1:\\ Input sentence: What a script, what a story, what a mess!\\ Sentiment: negative\\  Example 2:\\ Input sentence: This is a great movie. \\ Too bad it is not available on home video.\\ Sentiment: positive\\ \#\#\#\\ 1. {[}INPUT\_SENTENCE1{]}\\ 2. {[}INPUT\_SENTENCE2{]}\end{tabular} & \begin{tabular}[c]{@{}l@{}}\#\#\#\\ Read the sentences and identify their sentiment. \\ You should think step by step and output the sentiment of each sentence.\\ There are two optional sentiments: negative or positive\\  Example 1:\\ Input sentence: What a script, what a story, what a mess!\\ Think step by step: He/she thinks the script and story are messy. Therefore, the sentiment is negative.\\ Sentiment: negative\\Example 2:\\ Input sentence: This is a great movie. Too bad it is not available on home video.\\ Think step by step: He/she thinks the movie is great and he/she would like to watch it on home video. \\ He/she really love this movie. Therefore, the sentiment is positive\\ Sentiment: positive\\ \#\#\#\\  1. {[}INPUT\_SENTENCE1{]}\\ 2. {[}INPUT\_SENTENCE2{]}\end{tabular} \\ \midrule             
KGC & \begin{tabular}[c]{@{}l@{}}\#\#\#\\ Read the word pairs and identify their semantic relation. \\ You have the following options: entailment, hyponym, hypernym, ...\\ Here are some examples that you can refer to.\\ Example 1:\\ Input word pair: 3D, film\\ semantic relation: hyponym\\ Example 2:\\ Input word pair: ability, unfitness\\ semantic relation: antonym\\ Example 3:\\ Input word pair: abominably,atrociously\\ semantic relation: synonym\\ Example 4:\\ Input word pair: acclaim, sanction\\ semantic relation: entailment\\ ...\\ \#\#\#\\ 1. {[}INPUT\_WORD1{]}, {[}INPUT\_WORD2{]}\\ 2. {[}INPUT\_WORD3{]}, {[}INPUT\_WORD4{]}\\ ...\end{tabular} & \begin{tabular}[c]{@{}l@{}}\#\#\#\\ Read the word pairs and identify their semantic relation. \\ You should think step by step and output the semantic relation of each sentence.\\ You have the following options: hyponym, antonym, synonym, hypernym, entailment, member,...\\ Here are some examples that you can refer to.\\ Example 1:\\ Input word pair: 3D, film\\ Think step by step: 3D films are a type of film, so the relationship between the two words is hyponym\\ semantic relation: hyponym\\Example 2:\\ Input word pair: ability, unfitness\\ Think step by step: In most cases, the term "unfitness" refers to anything or someone \\ who is not appropriate for the task at hand, meaning that their ability is inadequate. \\ Therefore, the relationship between the two words is antonym\\ semantic relation: antonym\\ Example 3:\\ Input word pair: abominably,atrociously\\ Think step by step: Abhorrently and atrociously both refer to something that is extremely unpleasant or unfair. \\ Therefore, the relationship between the two words is synonym\\ semantic relation: synonym\\ Example 4:\\ Input word pair: acclaim, sanction\\ Think step by step: Acclaim is used to describe enthusiastic public praise. \\ The verb sanction can be used to signify to grant permission or approval for (an activity). \\ So the relationship between the two words is entailment\\ semantic relation: entailment\\ ...\\ \#\#\#\\ 1. {[}INPUT\_WORD1{]}, {[}INPUT\_WORD2{]}\\ 2. {[}INPUT\_WORD3{]}, {[}INPUT\_WORD4{]}\end{tabular} \\ \midrule

Tagging & \begin{tabular}[c]{@{}l@{}}\#\#\#\\ Read the following sentences, choose one or more tags from the TagSet \\ that correspond to each sentence's semantics.\\ Here are one example that you can refer to.\\ Example 1: Input sentence: the metabolic world of escherichia coli ...\\ Tag list: ...\\ TagSet=\{...\}\\ \#\#\#\\ 1. {[}INPUT\_SENTENCE1{]}\\ 2. {[}INPUT\_SENTENCE2{]}\end{tabular} & \begin{tabular}[c]{@{}l@{}}\#\#\#\\ Read the following sentences, choose one or more tags from the TagSet \\ that correspond to each sentence's semantics.\\ You should think step by step and output the tag list of each sentence.\\ Here are one example that you can refer to.\\ Example 1: Input sentence: the metabolic world of escherichia coli is not small to elucidate ...\\ Think step by step: ...\\ Tag list: ...\\ TagSet=\{...\}\\ \#\#\#\\ 1. {[}INPUT\_SENTENCE1{]}\\ 2. {[}INPUT\_SENTENCE2{]}\end{tabular} \\ \bottomrule
\end{tabular}%
}
\caption{ChatGPT prompts for three annotation tasks. We consider four kinds prompt, including zero-shot, few-shot, zero-shot CoT, and few-shot CoT prompts.}
\label{tab:my-chatgpt}
\end{table*}

\end{document}